\documentclass[10pt,twocolumn,letterpaper]{article}

\usepackage[pagenumbers]{cvpr} 

\definecolor{cvprblue}{rgb}{0.21,0.49,0.74}
\usepackage[pagebackref,breaklinks,colorlinks,allcolors=cvprblue]{hyperref}

\def\paperID{} 
\def\confName{}
\def\confYear{}

\usepackage[most]{tcolorbox}
\tcbuselibrary{theorems}
\usepackage{amsmath,amssymb}

\usepackage[most]{tcolorbox}
\usepackage{amsthm}
\definecolor{wine}{rgb}{0.5333333333333333, 0.13333333333333333, 0.3333333333333333}

\newtheorem{theorem}{Theorem}[section]
\newtheorem{corollary}{Corollary}[section]
\newtheorem{definition}{Definition}[section]
\newtheorem{proposition}{Proposition}[section]

\definecolor{deblue}{RGB}{11,132,147}
\usepackage{enumitem}
\usepackage{tikz}
\newcommand{\fcircle}[2][red,fill=red]{\tikz[baseline=-0.5ex]\draw[#1,radius=#2] (0,0.03) circle ;}

\usepackage{amsthm}

\newtheoremstyle{simpleRemark}
  {\topsep}   
  {\topsep}   
  {\normalfont}  
  {}          
  {\bfseries\itshape} 
  {.}         
  { }         
  {}          

\theoremstyle{simpleRemark}

\usepackage{amsmath}

\newcommand{\nocontentsline}[3]{}
\let\origcontentsline\addcontentsline
\newcommand\stoptoc{\let\addcontentsline\nocontentsline}
\newcommand\resumetoc{\let\addcontentsline\origcontentsline}

\usepackage{multirow}
\usepackage{graphicx}
\usepackage{makecell}
\usepackage{xcolor}
\usepackage{colortbl}
\usepackage{hhline}
\usepackage{arydshln}

\usepackage{cuted}
\usepackage[misc]{ifsym}

\title{Deep Spectral Prior}

\author{\hspace{-0.3cm}\textbf{Yanqi Cheng$^1$ Xuxiang Zhao$^2$ Tieyong Zeng$^3$ Pietro Lio$^1$  Carola-Bibiane Schönlieb$^1$} \\[0.1cm]
\textbf{Angelica I Aviles-Rivero}$^{4}$\Letter \\[0.3cm]
  $^1$ University of Cambridge
  $^2$ Qiuzhen College, Tsinghua University\\
  $^3$ The Chinese University of Hong Kong
  $^4$ YMSC, Tsinghua University
}

\begin{document}
\stoptoc

\maketitle

\begin{strip}
\begin{minipage}{\textwidth}\centering
\vspace{-30pt}
\includegraphics[width=\textwidth]{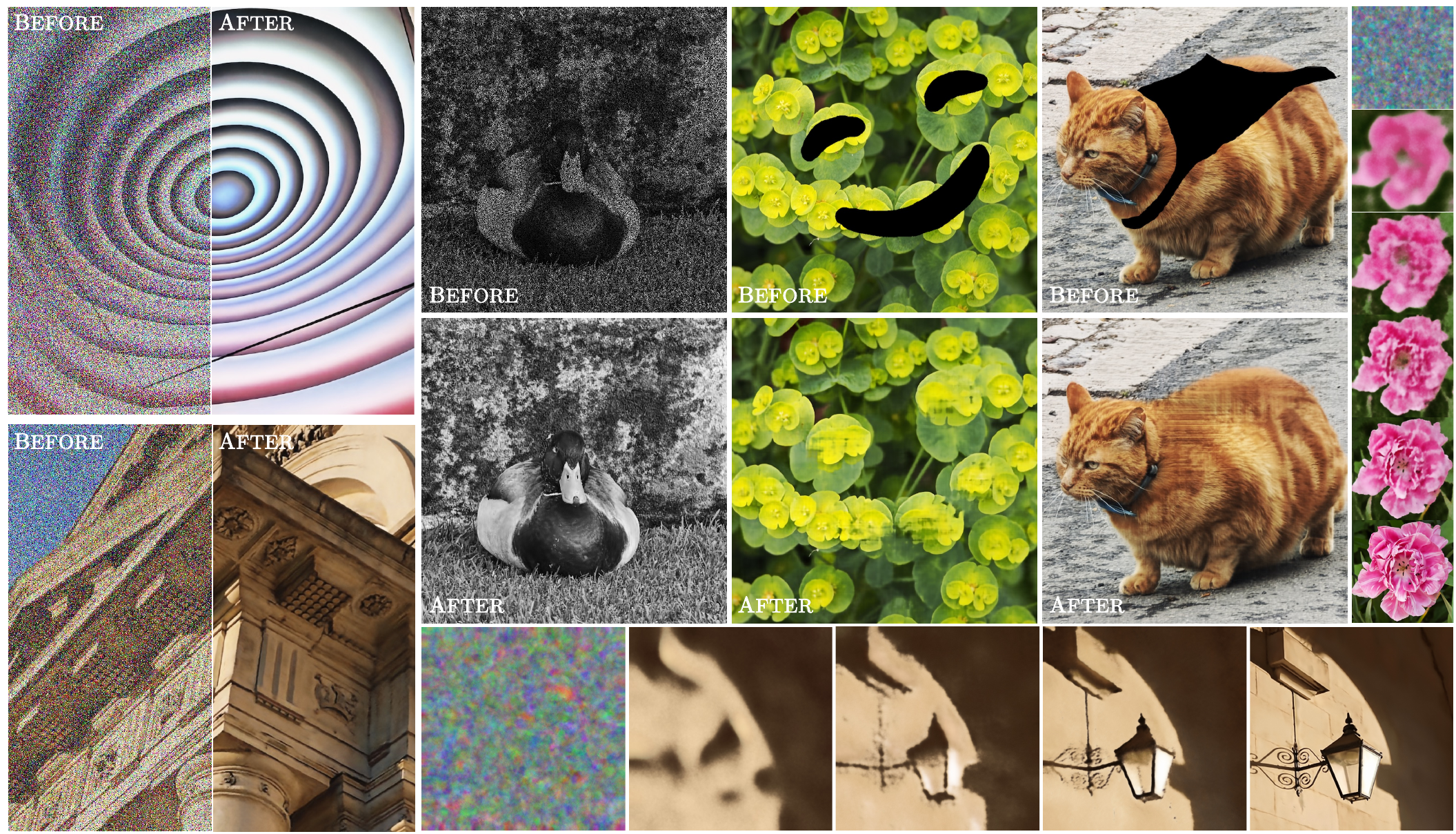}
\label{figurelabel}
\vspace{-10pt}
\end{minipage}
\end{strip}

\begin{abstract}
We introduce the Deep Spectral Prior (DSP), a new framework for unsupervised image reconstruction that operates entirely in the complex frequency domain. Unlike the Deep Image Prior (DIP), which optimises pixel-level errors and is highly sensitive to overfitting, DSP performs  joint learning of amplitude and phase to capture the full spectral structure of images.
We derive a rigorous theoretical characterisation of DSP’s optimisation dynamics, proving that it follows frequency-dependent descent trajectories that separate informative low-frequency modes from stochastic high-frequency noise. This spectral mode separation explains DSP’s self-regularising behaviour and, for the first time, formally establishes the elimination of DIP’s major limitation—its reliance on manual early stopping. Moreover, DSP induces an implicit projection onto a frequency-consistent manifold, ensuring convergence to stable, physically plausible reconstructions without explicit priors or supervision.
Extensive experiments on denoising, inpainting, and deblurring demonstrate that DSP consistently surpasses DIP and other unsupervised baselines, achieving superior fidelity, robustness, and theoretical interpretability within a unified, unsupervised data-free framework.
\end{abstract}

\section{Introduction}
Inverse problems lie at the heart of image restoration and reconstruction, encompassing tasks such as denoising, inpainting, super-resolution, and deblurring. Traditionally approached through handcrafted priors and variational optimisation~\cite{rudin1992nonlinear,chan2001active}, these problems have seen a paradigm shift with the advent of deep learning, with single-image techniques effectively leverage data~\cite{essakine2025where,cheng2024singleshot,cheng2024you}.
Among the most intriguing developments in this space is the Deep Image Prior (DIP)~\cite{ulyanov2018deep}, which showed that the structure of an untrained convolutional neural network (CNN) alone can act as a powerful prior. By optimising a randomly initialised network to match a single degraded observation, DIP enables high-quality reconstructions—without requiring any training data. 
This surprising result has made DIP a widely adopted framework across diverse domains, including hyperspectral~\cite{rasti2021undip,li2025enhanced}, microscopy~\cite{vu2021deep, zhou2020diffraction}, compressed sensing~\cite{van2018compressed}, and medical imaging~\cite{gong2018pet,yoo2021time,barbano2022educated,liang2025analysis,liang2025sequential}.

Despite its empirical success, DIP suffers from two key limitations that hinder its practical effectiveness. First, DIP is notoriously prone to overfitting. Due to the lack of explicit regularisation~\cite{rudin1992nonlinear, liu2019image, chang2000adaptive,chambolle2004algorithm,getreuer2012rudin}, the network can memorise noise or artifacts if optimisation proceeds too long, requiring ad hoc early stopping based on visual inspection~\cite{wang2021early}. Second, DIP’s pixel-wise reconstruction loss provides no explicit mechanism to control the frequency content or semantic structure of the recovered image. As a result, it tends to favor smooth, low-frequency components early in optimisation (due to spectral bias~\cite{rahaman2019spectral}), while high-frequency details—such as texture or fine edges—are either fit too late or overfit entirely if optimisation proceeds unchecked~\cite{rahaman2019spectral, ulyanov2018deep}.

To mitigate these issues, recent works have explored two main directions~\cite{cheng2019bayesian, jo2021rethinking, chen2020dip, cascarano2022first, zhang2025survey}. One line of research modifies the DIP architecture or optimisation dynamics—e.g., by freezing weights and optimising the input~\cite{sun2021plug}. 
Another line of work augments DIP with explicit regularisers, such as Total Variation (TV) and variants~\cite{rudin1992nonlinear, liu2019image, chambolle2004algorithm,getreuer2012rudin} or Regularisation by Denoising (RED)~\cite{romano2017little, mataev2019deepred}. From the perspective of inverse problems, such \textit{double-prior formulations} are theoretically beneficial~\cite{arndt2022regularization}, as additional constraints help mitigate the inherent ill-posedness. However, these approaches fundamentally alter the nature of DIP by introducing explicit handcrafted or learned priors, additional hyperparameters, and tuning requirements. They move away from DIP’s core appeal: solving inverse problems using only the implicit regularisation of a randomly-initialised network, without any external data or assumptions. 
Once explicit regularisation is introduced, it becomes unclear whether performance gains stem from the network’s inductive bias or the added constraint. As such, direct comparisons with these methods obscure the role of implicit regularisation alone in reconstruction performance. In contrast, we propose the Deep Spectral Prior (DSP)—a novel spectral formulation that preserves its data-free nature while providing theoretical guarantees for stability and convergence.

\textbf{Contributions.} We introduce Deep Spectral Prior (DSP), a new unsupervised framework for solving inverse problems in imaging that operates directly in the frequency domain. In particular:

\fcircle[fill=deblue]{2pt}  \textit{A new frequency-domain formulation.}   We introduce inverse imaging as spectral alignment in the Fourier domain, where optimisation operates directly on the amplitude and phase components of the signal. 
This formulation (\textit{Definition~1}) enables explicit control over frequency modes and corrects the low-frequency bias inherent in DIP, yielding an intrinsic spectral inductive bias aligned with both image statistics and neural spectral dynamics.

\fcircle[fill=deblue]{2pt} We provide, for the first time, a rigorous analysis of the optimisation dynamics of the Deep Spectral Prior (DSP).  
This analysis characterises DSP’s implicit spectral regularisation through the following key results:\\
--  \textbf{Spectral equivalence:} We prove that the DSP loss is mathematically equivalent to the pixel-domain reconstruction error under a unitary Fourier transform (Corollary~\ref{cor::2.1} ), preserving the reconstruction objective while reformulating the optimisation landscape. \\
-- \textbf{Complex descent dynamics:} We establish that DSP and DIP follow fundamentally different optimisation trajectories due to their underlying complex-valued descent laws (Theorem~\ref{thm::2.1}), providing a theoretical basis for DSP’s smoother and more stable convergence. \\
-- \textbf{Spectral stability and early-stopping robustness:} We show that  our Deep Spectral Prior exhibits frequency-ordered convergence—recovering low-frequency components first while naturally suppressing high-frequency noise—thereby eliminating the need for early stopping (see Theorem~\ref{thm::2.2} and Proposition~\ref{prop::2.1}). 

\fcircle[fill=deblue]{2pt} We evaluate DSP on several inverse problems, comparing against DIP, and recent deep unsupervised baselines. DSP consistently outperforms DIP across all tasks and images—often by significant margins.
Given the widespread and enduring use of DIP as a general-purpose implicit prior, we believe DSP opens a new line of research in frequency-domain implicit methods with similar long-term impact.

\section{Methodology: Deep Spectral Prior}
Conventional pixel-space approaches such as the Deep Image Prior (DIP) optimise over spatial intensities, neglecting the spectral organisation that fundamentally governs natural images. We introduce the Deep Spectral Prior (DSP)—the first principled framework that reformulates image reconstruction entirely in the frequency domain and provides a theoretical understanding of its dynamics. By operating on complex-valued representations, DSP aligns learning with the intrinsic spectral bias of neural networks, offering provable stability, noise robustness, and self-regularising behaviour absent in spatial formulations.

\subsection{From Pixels to Frequencies: Deep Spectral Prior}
Let the observed degraded image be \( y = \mathcal{A} x + \epsilon \), 
where \( \mathcal{A} \) is a known linear degradation operator and \( \epsilon \) denotes additive noise. 
Existing reconstruction approaches, including DIP, optimise pixel-wise fidelity but overlook global spectral coherence.
Yet  neural representations are inherently spectral, 
dominated by smooth low-frequency components and rapidly decaying high-frequency content.

\begin{figure}[t!]
  \centering
 \includegraphics[width=0.45\textwidth]{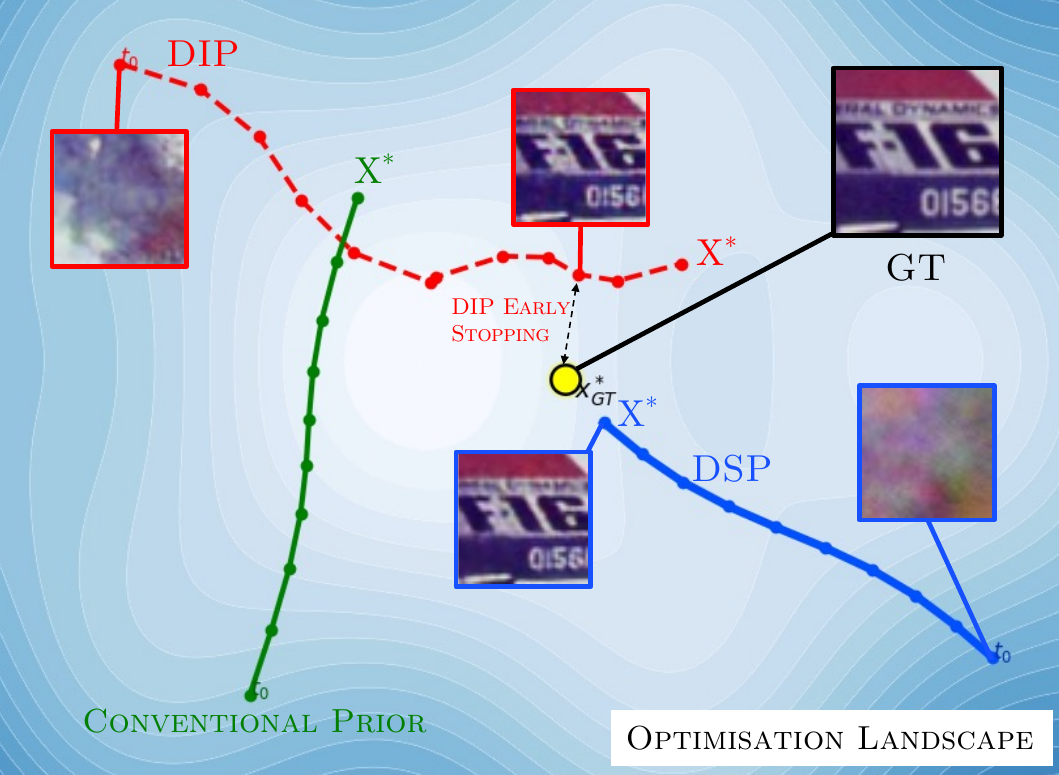}\\[-0.2cm]
  \caption{Optimisation trajectories for DIP, DSP, and  conventional priors.
DIP (red) overshoots and needs manual early stopping; the conventional prior (green) converges to a biased solution; DSP (blue) follows a stable spectral path and converges close to the GT without early stopping.
  }
\label{opt_landscape}
\end{figure}

This motivates a reformulation in the Fourier domain, 
where optimisation aligns frequency content rather than pixel intensities. 
Such a viewpoint naturally separates structured signal from stochastic noise 
and introduces frequency-dependent inductive bias with explicit control over spectral contributions. 
We next introduce the \emph{Deep Spectral Prior (DSP)}, 
a framework that performs reconstruction directly in the frequency domain, 
revealing a new form of implicit spectral regularisation (see Figure~\ref{opt_landscape}).

\begin{definition}{\textbf{Deep Spectral Prior (DSP).}}\label{def:DSP}
The \emph{Deep Spectral Prior (DSP)} formulates image reconstruction directly in the frequency domain using complex-valued optimisation. 
Let \(f_\theta(z)\in\mathbb{C}^{n\times n}\) denote the output of a neural network parameterised by  weights \(\theta\) 
for a fixed latent input \(z\in\mathbb{C}^d\), 
and let \(\mathcal{A}:\mathbb{R}^{n\times n}\!\to\!\mathbb{R}^{n\times n}\) 
be a known linear degradation operator acting on real-valued images. 
Given the observed corrupted image \(y\in\mathbb{R}^{n\times n}\), 
the DSP loss is defined as
\begin{equation}
\begin{aligned}
    \mathcal{L}_{\mathrm{DSP}}(\theta)
&=
\frac{1}{2}\big\|
\mathcal{F}(\mathcal{A} f_\theta(z))
-
\mathcal{F}(y)
\big\|_2^2,\\
\theta^* &= \arg\min_\theta \mathcal{L}_{\mathrm{DSP}}(\theta),
\end{aligned}
\end{equation}

\noindent where \(\mathcal{F}:\mathbb{R}^{n\times n}\!\to\!\mathbb{C}^{n\times n}\) 
is the discrete Fourier transform (DFT). 
All optimisation is performed in the complex domain, enabling the model to jointly learn amplitude and phase information  and to capture the spectral correlations.
\end{definition}

\subsection{Spectral Equivalence and the Complex Descent Law of DSP}
The Deep Spectral Prior (DSP) can be interpreted as a reformulation of image reconstruction in the frequency domain rather than the pixel domain.
Although this shift might appear purely representational, it fundamentally alters the geometry of optimisation.
We begin by establishing that the DSP loss is mathematically equivalent to the conventional pixel-space error, thereby preserving the reconstruction objective while transforming the optimisation landscape.
\begin{corollary}[\textbf{Equivalence of Deep Spectral Prior Loss and Pixel-Space Error}]~\label{cor::2.1}
Let \(x, \hat{x} \in \mathbb{R}^m\), and let \(\mathcal{F} : \mathbb{R}^m \to \mathbb{C}^m\) denote the unitary discrete Fourier transform. Then
\[
\|x - \hat{x}\|_2^2 = \|\mathcal{F}(x) - \mathcal{F}(\hat{x})\|_2^2.
\]
In particular, for any neural network output \(f_\theta(z)\) and observation \(y\),
$
\mathcal{L}_{\mathrm{DSP}}(\theta)
=\frac{1}{2}\big\|
\mathcal{F}(\mathcal{A} f_\theta(z)) - \mathcal{F}(y)
\big\|_2^2
$
is exactly equal to the pixel-domain reconstruction error \(\frac{1}{2}\|\mathcal{A}f_\theta(z) - y\|_2^2\). 
Thus, DSP does not alter the objective value but reformulates the optimisation in the spectral domain. (See Proof in the Supplementary Material)
\end{corollary}

While the loss values in pixel and Fourier domains are equivalent under a unitary transform, 
the optimisation trajectories are not. 
In particular, real-valued networks, including that of DIP, and complex-valued networks (DSP) follow fundamentally different descent dynamics, 
as formalised below.

\begin{theorem}{\textnormal{\textbf{(Complex Descent Law: Distinct Optimisation Paths of DIP and DSP).}}}{}~\label{thm::2.1}
Consider the frequency-domain loss
\begin{align}
\mathcal{L}_{\mathrm{DSP}}(\theta)
&=
\frac{1}{2}\big\|
\mathcal{F}(\mathcal{A}f_\theta(z)) - \mathcal{F}(y)
\big\|_2^2\label{eq:L2}\\
&=
\frac{1}{2}
\big(
\mathcal{F}(\mathcal{A}f_\theta(z)) - \mathcal{F}(y)
\big)^\dagger
\big(
\mathcal{F}(\mathcal{A}f_\theta(z)) - \mathcal{F}(y)
\big).\nonumber
\end{align}
where \(\mathcal{F}\) denotes the discrete Fourier transform and \({}^\dagger\) the Hermitian transpose.
The optimisation dynamics under gradient descent differ fundamentally between real-valued DIP and complex-valued DSP:

\begin{itemize}
\item \textbf{DIP Optimisation:} for \(\theta \in \mathbb{R}^p\),
\begin{align}
&    \frac{d\theta}{dt}
=
-\alpha\nabla_\theta \mathcal{L}_{\mathrm{DIP}}(\theta)\\
&=
-\alpha\,
\nabla_\theta
\big(
\mathcal{A}f_\theta(z)- y
\big)^T
\!\cdot\!
\big(
\mathcal{A}f_\theta(z) - y
\big)
,\nonumber\\
&
\text{where} \,\, \mathcal{L}_{\mathrm{DIP}}(\theta)
=
\frac{1}{2}\big\|
\mathcal{A} f_\theta(z)
-
y
\big\|_2^2\nonumber.
\end{align}

\item \textbf{DSP Optimisation:} for \(\tilde{\theta} = a + ib \in \mathbb{C}^p\),
\begin{align}
&\frac{da}{dt} = -\alpha\,\nabla_a \mathcal{L}_{\mathrm{DSP}}(\tilde{\theta}),
\frac{db}{dt} = -\alpha\,\nabla_b \mathcal{L}_{\mathrm{DSP}}(\tilde{\theta}), \nonumber\\[4pt]
&\frac{d\tilde{\theta}}{dt} = -\alpha\,\nabla_{\tilde{\theta}^*} \mathcal{L}_{\mathrm{DSP}}(\tilde{\theta}) \label{eq:dsp_dynamics}\\[4pt]
&= -\alpha\,
\nabla_{\tilde{\theta}^*}
\Big[
\big(
\mathcal{F}(\mathcal{A}f_{\tilde{\theta}}(z)) - \mathcal{F}(y)
\big)^\dagger
\!\cdot\!
\big(
\mathcal{F}(\mathcal{A}f_{\tilde{\theta}}(z)) - \mathcal{F}(y)
\big)
\Big].\nonumber
\end{align}
where \(\nabla_{\tilde{\theta}^*}\) denotes the Wirtinger gradient with respect to the complex conjugate variable. (See Proof in the Supplementary Material)
\end{itemize}
\end{theorem}

We undeline that our DSP formulation also applies when the network parameters are real-valued, 
since the real gradient corresponds to the real part of the Wirtinger derivative. 
Hence, DSP can be implemented with standard real-valued backbones, 
as the induced directional decomposition still captures the distinct curvature 
and convergence geometry of the spectral optimisation landscape.

\subsection{The Spectral Stability Law: Robust Early-Stopping Dynamics of DSP}

We next show that the complex-valued Deep Spectral Prior (DSP) naturally provides a more stable and extended early stopping window than the conventional  Deep Image Prior (DIP).  
This property can be interpreted through  the {Neural Tangent Kernel (NTK)}~\cite{jacot2018neural} framework, revealing that DSP induces smoother optimisation dynamics and mitigates premature overfitting.

\begin{theorem}{\textnormal{\textbf{(Robust Stopping Property of  DSP).}}}{}~\label{thm::2.2}
For a DSP optimisation trajectory, the {NTK approximation} takes the form
\begin{equation}\label{NTK_approx_spectral}
\mathcal{F}(\mathcal A f_{\tilde\theta}(z))
=
\mathcal{F}(\mathcal Af_{\tilde{\theta}_0}(z)) + \tilde{\Phi}(z)\tilde{u},
\end{equation}
where $\tilde{\Phi}(z)=\frac{\partial \mathcal F(\mathcal{A}f_{\tilde{\theta}}(z))}{\partial \tilde\theta}\in\mathbb{C}^{N\times p}$ is assumed to be a constant over steps, with $N = H\times W\times3$, $p=\dim(\tilde \theta)$, and parameters without initialisation $\tilde u=\tilde\theta-\tilde\theta_0$ . 
DSP optimisation path (\ref{eq:dsp_dynamics}) guides parameters $\tilde u$ to update along the path:
\begin{equation}\label{eq:DSP-parameter}
\frac{d\tilde u}{dt}=
-\alpha\,\tilde{\Phi}(z)^\dagger(\tilde{\Phi}(z)\tilde{u} - \tilde{y}),
\end{equation}
where $\alpha$ is the gradient descent step size. 
The evolution of the discrepancy along gradient descent follows a linear ordinary differential equation (ODE) governed by the Hermitian operator $\tilde{\Phi}\tilde{\Phi}^\dagger$:
\begin{equation}\label{eq:DSP-picture}
\frac{d(\tilde{\Phi}(z)\tilde{u} - \tilde{y})}{dt}
=
-\alpha\,\tilde{\Phi}(z)\tilde{\Phi}(z)^\dagger(\tilde{\Phi}(z)\tilde{u} - \tilde{y}),
\end{equation} 
The ODE admits the closed-form solution
\begin{equation}\label{eq:DSP_solution}
\mathcal{F}(\mathcal A f_{\tilde\theta^{(t)}}(z))
=
\sum_{k=1}^{N}(1 - e^{-\alpha\mu_k t})
\langle \tilde{y}, \tilde{\psi}_k\rangle \tilde{\psi}_k,
\end{equation}
where $\tilde{\psi}_k$ and $\mu_k$ satisfy the eigendecomposition
$\tilde{\Phi}\tilde{\Phi}^\dagger
=
\sum_{k=1}^N \mu_k \tilde{\psi}_k \tilde{\psi}_k^\dagger$.
The mean-squared error (MSE) along the optimisation path of DSP evolves as
\begin{align}
&MSE_t
= \mathbb{E}\Vert \mathcal{F}(\mathcal A f_{\tilde{\theta}^{(t)}}(z)) - \tilde x\Vert^2 \label{eq:mse_t}\\
&= \sum_{k=1}^{N}\!\Big[
 (1 - e^{-\alpha\mu_k t})^2(\mathbb{E}|\langle \tilde{\epsilon}, \tilde{\psi}_k\rangle|^2+|\langle\tilde{\mathcal A x},\tilde\psi_k\rangle|^2)\nonumber\\
 & -
 2(1 - e^{-\alpha\mu_k t})Re\big
 \{\langle\tilde{\mathcal Ax},  \tilde\psi_k\rangle^*\langle\tilde x,  \tilde\psi_k\rangle\big\}+|\langle \tilde x,\tilde \psi_k \rangle|^2
\Big]\!. \nonumber
\end{align}
where $\tilde{x}=\mathcal F(x)$, $\tilde{\mathcal A x}=\mathcal F(\mathcal {A}x)$ and noise $\tilde \epsilon=\mathcal F(\epsilon)$.
The optimal stopping time for each mode $k$ is then
\vspace{-0.2cm}
\begin{align}
t_k^* &= \frac{1}{\alpha\mu_k}\ln\!\left(\rho_k\right), \label{eq:DSP_stopping} \\
\rho_k &=
\frac{\mathbb{E}|\langle \tilde{\epsilon}, \tilde{\psi}_k\rangle|^2+|\langle\tilde{\mathcal A x},\tilde\psi_k\rangle|^2}
{\mathbb{E}|\langle \tilde{\epsilon}, \tilde{\psi}_k\rangle|^2+|\langle\tilde{\mathcal A x},\tilde\psi_k\rangle|^2
- Re\{\langle\tilde {\mathcal{A}x},\tilde\psi_k\rangle^*\langle\tilde x,\tilde \psi_k\rangle\}}. \notag
\end{align}

\noindent Different eigenvalues $\mu_k$ induce distinct convergence speeds, endowing DSP with an inherent robustness to  stopping time. (See proof in the Supplementary Material)
\end{theorem}
\noindent\textbf{Sketch of Proof.}  
The NTK form in \eqref{NTK_approx_spectral} arises from a first-order Taylor expansion of $\mathcal{F}(\mathcal{A}f_{\tilde{\theta}}(z))$ around $\tilde{\theta}_0$, assuming $\tilde{\Phi}(z)$ remains approximately constant in early training. 
This leads directly to the parameter-space ODE (\ref{eq:DSP-parameter}). 
Multiplying both sides by $\tilde{\Phi}(z)$ gives the discrepancy dynamics (\ref{eq:DSP-picture}). 
Because $\tilde{\Phi}\tilde{\Phi}^\dagger$ is Hermitian and positive semidefinite, it admits the eigendecomposition $\tilde{\Phi}\tilde{\Phi}^\dagger = \sum_k \mu_k \tilde{\psi}_k \tilde{\psi}_k^\dagger$, which decouples the ODE into $N$ independent spectral components. 
Each mode evolves as $\frac{d\langle\tilde{\Phi}\tilde{u}-\tilde{y},\tilde{\psi}_k\rangle}{dt}=-\alpha\mu_k\langle\tilde{\Phi}\tilde{u}-\tilde{y},\tilde{\psi}_k\rangle$, yielding the closed-form solution (\ref{eq:DSP_solution}). 

Substituting into the MSE definition and using $\tilde{y}=\tilde{\mathcal{A}x}+\tilde{\epsilon}$ with $\mathbb{E}[\tilde{\epsilon}]=0$ removes cross-terms, resulting in (\ref{eq:mse_t}). 
Differentiating $\mathrm{MSE}_t$ and solving $\frac{d\mathrm{MSE}_t}{dt}=0$ produces (\ref{eq:DSP_stopping}). 
Large $\mu_k$ correspond to smooth, low-frequency modes that converge rapidly, while small $\mu_k$ represent noise-dominated components that decay slowly. 
This spectral separation ensures stable convergence and naturally eliminates the need for explicit early stopping. (See full Proof in the Supplementary Material). \hfill$\square$

\begin{figure*}[t!]
  \centering
\includegraphics[width=\textwidth]{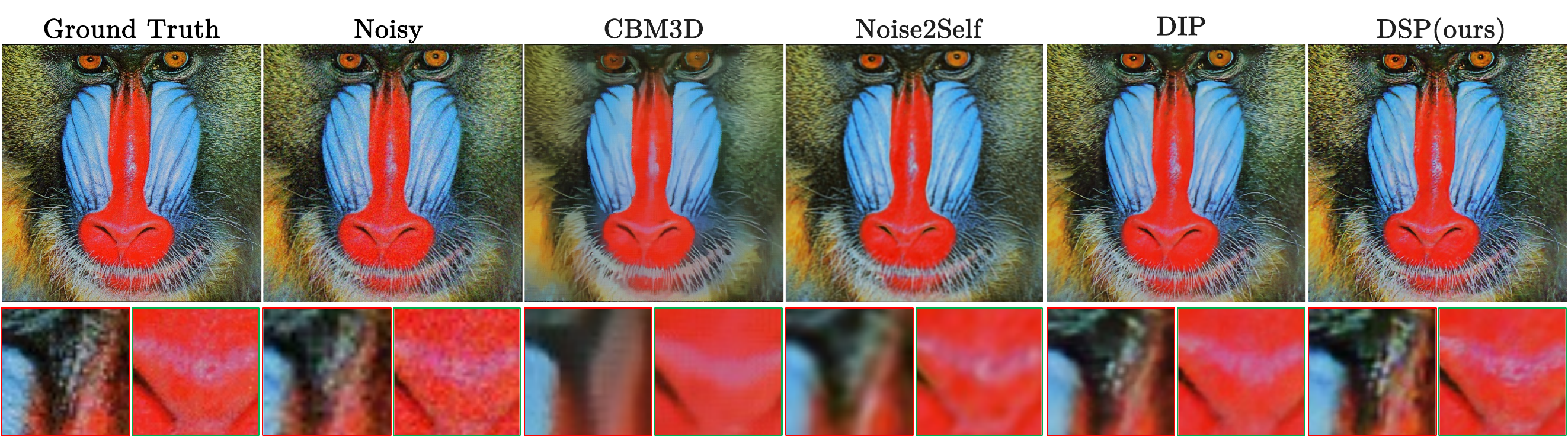}\\[-0.2cm]
  \caption{Visual comparison of blind denoising on the “Baboon” image. DSP (ours) is compared vs. unsupervised baselines, with zoomed-in views highlighting fine details preservation.
  }
\label{denoising_baboon}
\vspace{-0.3cm}
\end{figure*}
\begin{figure*}[t!]
  \centering
\includegraphics[width=\textwidth]{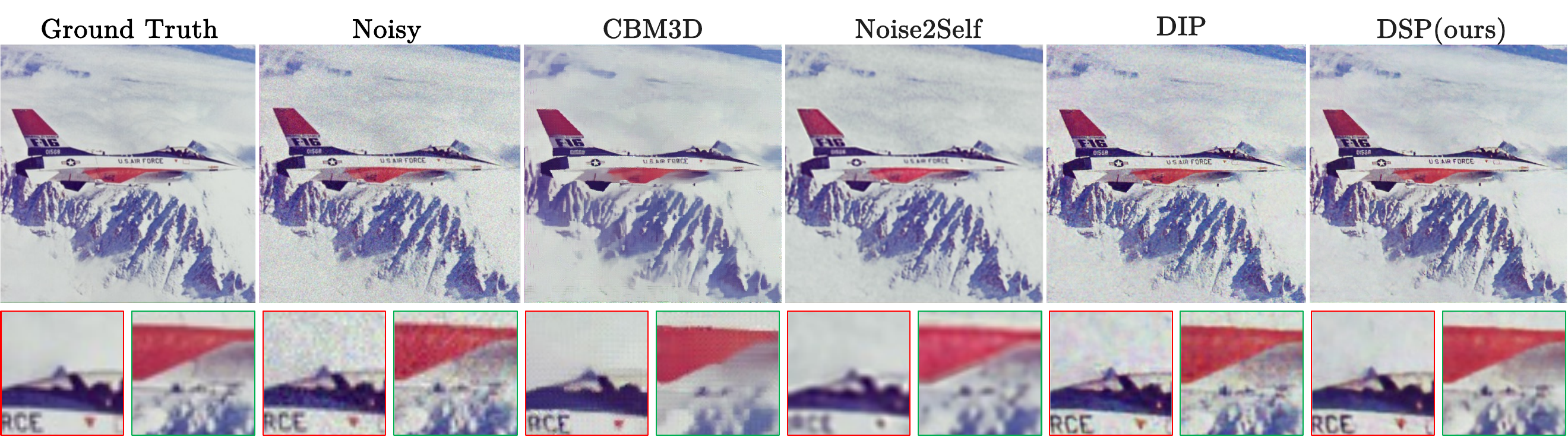}\\[-0.2cm]
  \caption{Blind denoising comparison on the ‘Plane’ image: our DSP method vs. unsupervised baselines, with zoomed-in views highlighting the retention of fine details.
  }
\label{denoising_plane}
\vspace{-0.3cm}
\end{figure*}
\begin{figure*}[t!]
  \centering
\includegraphics[width=\textwidth]{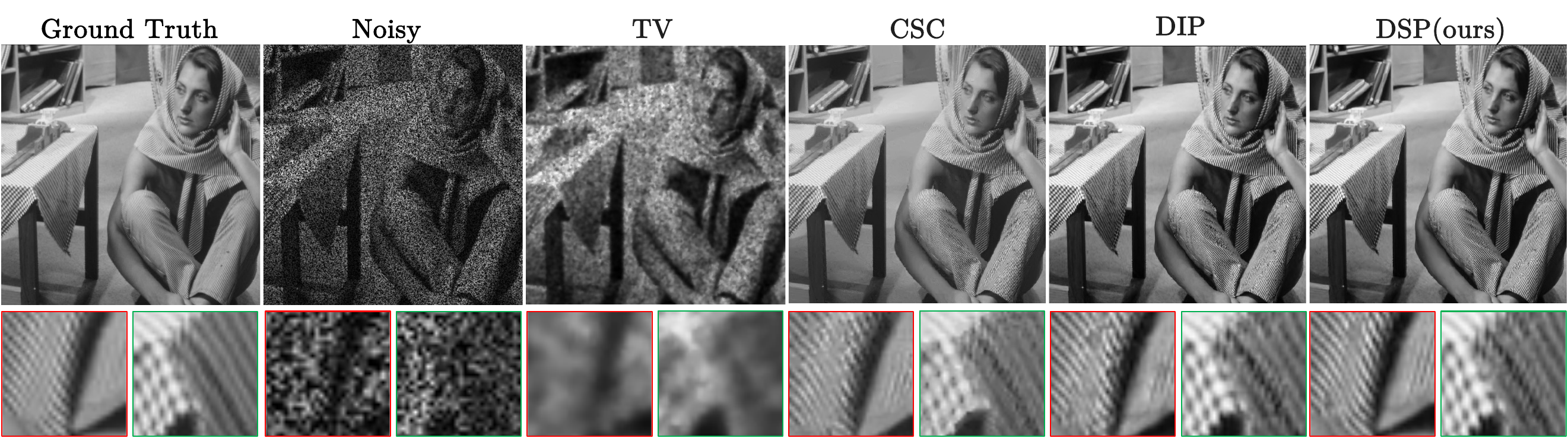}\\[-0.2cm]
  \caption{Restoration results on the “Barbara” image with heavy noise. DSP (ours) better preserves structured textures (e.g., fabric) compared to DIP, TV, and CSC, as shown in the zoomed-in regions.}
  \label{restoration_barbara} 
  \vspace{-0.3cm}
\end{figure*}

The expression for $t_k^*$ reveals that modes associated with larger eigenvalues $\mu_k$ converge faster and capture the structured, low-frequency content of the image, while modes with small $\mu_k$ correspond to high-frequency noise. This spectral separation induces a  stabilisation of optimisation: informative components settle early, whereas noise remains suppressed. Consequently, \textit{DSP self-regularises the training dynamics, removing the need for explicit early stopping}. A higher-dimensional latent space further accentuates this frequency separation, enhancing convergence robustness.

This result provides, to our knowledge, the first theoretical justification showing that a Fourier-domain formulation---as introduced in DSP---inherently mitigates the need for early stopping. By optimising directly in the frequency domain, DSP prioritises low-frequency fidelity while attenuating slow-evolving high-frequency components, leading to stable convergence without manual intervention.
Having established the asymptotic convergence behaviour, we next examine how this frequency selectivity emerges \textit{during} training. In particular, we analyse the evolution of reconstruction in the spectral domain to characterise the implicit frequency ordering of DSP optimisation.

\begin{proposition}[\textbf{Gradient Dynamics in Frequency Space}]~\label{prop::2.1}
Consider the frequency-domain loss
$
\mathcal{L}_{\mathrm{DSP}}(\theta) := \left\| \mathcal{F}(\mathcal{A}f_\theta(z)) - \mathcal{F}(y) \right\|_2^2.
$
Then, when the parameters $\theta$ are updated by a smooth optimisation algorithm with small learning rate, the iterates $\{\theta_t\}$ yield reconstructions $u_t := \mathcal{A}f_{\theta_t}(z)$ such that
\[
\mathcal{F}(u_t) \to \mathcal{F}(y) \quad \text{in } \mathbb{C}^m.
\]
Moreover, if the network $f_\theta$ exhibits spectral bias, then the components of $\mathcal{F}(u_t)$ at low spatial frequencies converge faster than those at high frequencies. 
\end{proposition}

\begin{table*}[t!]
\caption{Performance comparison (PSNR in dB) across unsupervised denoising methods (DIP, DSP, CBM3D, Noise2Self, Wavelet, Total Variation (TV), TV with split Bregman, WIRE and DS-N2N) on various images. Best values per image are highlighted in bold.}
\label{psnr_comparison_denoise}
\setlength{\tabcolsep}{4pt}
\vspace{-0.2cm}
\resizebox{\linewidth}{!}{
\hspace{0.025\linewidth}
\begin{tabular}{
    >{\centering\arraybackslash}p{2.5cm}|
    >{\centering\arraybackslash}p{1cm}|
    >{\centering\arraybackslash}p{1.8cm}
    >{\centering\arraybackslash}p{2.1cm}
    >{\centering\arraybackslash}p{1.75cm}
    >{\centering\arraybackslash}p{1.95cm}
    >{\centering\arraybackslash}p{1.75cm}
    >{\centering\arraybackslash}p{1.75cm}
    >{\centering\arraybackslash}p{1.65cm}
}
\Xhline{0.25ex}
\multicolumn{2}{c|}{\textsc{Method}} & Glass Roof & \hspace{-0.3cm}Orange Building & \hspace{-0.3cm}Brick Wall & \hspace{-0.2cm}Brick Building & Baboon & Monarch & Plane \\
\hline
Wavelet~\cite{chang2000adaptive}&2000 & 21.37 & 26.54 & 24.76 & 22.19 & 22.16 & 28.91 & 28.45 \\
TV~\cite{chambolle2004algorithm}&2004 & 23.90 & 23.55 & 23.90 & 23.40 & 23.29 & 25.04 & 28.70 \\
CBM3D~\cite{lebrun2012analysis}&2012 & 19.73 & 21.95 & 20.57 & 19.86 & 20.77 & 26.28 & 27.29 \\
TV Bregman~\cite{getreuer2012rudin}&2012 & 17.45 & 22.70 & 21.29 & 18.58 & 19.53 & 23.26 & 24.28 \\
DIP~\cite{ulyanov2018deep}&2018 & 24.26 & 27.65 & 24.27 & 22.95 & 22.77 & 30.99 & 28.83 \\
Noise2Self~\cite{batson2019noise2self}&2019 & 22.81 & 25.19 & 23.81 & 21.94 & 21.31 & 28.92 & 28.64 \\
WIRE~\cite{saragadam2023wire}&2023 & 19.92 & 25.41 & 22.93 & 20.35 & 21.57 & 29.22 & 29.64 \\
DS-N2N~\cite{bai2025dual}&2025 &25.45&27.96&25.77&23.56&23.36&30.65&30.40\\
\cellcolor[HTML]{D7FFD7}DSP (ours) &
\cellcolor[HTML]{D7FFD7}2025& 
\cellcolor[HTML]{D7FFD7}\textbf{25.59} & 
\cellcolor[HTML]{D7FFD7}\textbf{28.50} & 
\cellcolor[HTML]{D7FFD7}\textbf{25.80} & 
\cellcolor[HTML]{D7FFD7}\textbf{23.85} & 
\cellcolor[HTML]{D7FFD7}\textbf{23.59} & 
\cellcolor[HTML]{D7FFD7}\textbf{31.13} & 
\cellcolor[HTML]{D7FFD7}\textbf{30.52} \\
\Xhline{0.25ex}
\end{tabular}
}
\vspace{-0.2cm}
\end{table*}

\begin{table*}[t!]
\caption{Performance comparison (PSNR in dB) across learned methods (CSC, DIP, DSP (ours)) on various images in Set12, with restoration tasks. Best values per image are highlighted in bold.}
\label{psnr_comparison_restoration}
\renewcommand{\arraystretch}{1.1}
\setlength{\tabcolsep}{4pt}
\vspace{-0.2cm}
\resizebox{\linewidth}{!}{
\hspace{0.025\linewidth}
\begin{tabular}{
    >{\centering\arraybackslash}p{2.55cm}|
    >{\centering\arraybackslash}p{1cm}|
    >{\centering\arraybackslash}p{1.85cm}
    >{\centering\arraybackslash}p{1.85cm}
    >{\centering\arraybackslash}p{1.85cm}
    >{\centering\arraybackslash}p{1.85cm}
    >{\centering\arraybackslash}p{1.85cm}
    >{\centering\arraybackslash}p{1.85cm}
    >{\centering\arraybackslash}p{1.85cm}
}
\Xhline{0.25ex}
\multicolumn{2}{c|}{\textsc{Method}}   & Photo & House & Pepper & Starfish & Parrot & Man & Barbara \\
\hline
CSC~\cite{wohlberg2016boundary}&2016  & 28.22 & 33.68 & 28.22  & 28.88    & 28.02  & 31.64 & 29.18 \\
DIP~\cite{ulyanov2018deep} &2018      & 29.61 & 37.22 & 32.68  & 32.55    & 29.49  & 31.83 & 30.97 \\
\cellcolor[HTML]{D7FFD7}DSP (ours) & 
\cellcolor[HTML]{D7FFD7}2025 &
\cellcolor[HTML]{D7FFD7}\textbf{29.74} & 
\cellcolor[HTML]{D7FFD7}\textbf{38.05} & 
\cellcolor[HTML]{D7FFD7}\textbf{32.81} & 
\cellcolor[HTML]{D7FFD7}\textbf{32.95} & 
\cellcolor[HTML]{D7FFD7}\textbf{29.60} & 
\cellcolor[HTML]{D7FFD7}\textbf{32.29} & 
\cellcolor[HTML]{D7FFD7}\textbf{32.33} \\
\Xhline{0.25ex}
\end{tabular}
}
\vspace{-0.1cm}
\end{table*}
\begin{table*}[t!]
\caption{PSNR (dB) comparison across super-resolution methods (Bicubic, DIP, TV, LapSRN, ResShift, SinSR (with single step and 15 steps in~\cite{wang2024sinsr}), and ASID with our proposed DSP on various test images. Best values for both supervised and unsupervised groups are highlighted in bold.}
\label{psnr_comparison_sr}
\renewcommand{\arraystretch}{1.1}
\setlength{\tabcolsep}{4pt}
\vspace{-0.2cm}
\resizebox{\linewidth}{!}{
\hspace{0.025\linewidth}
\begin{tabular}{
    >{\centering\arraybackslash}p{0.3cm} 
    >{\centering\arraybackslash}p{2cm}|
    >{\centering\arraybackslash}p{1cm}|
    >{\centering\arraybackslash}p{1.17cm}
    >{\centering\arraybackslash}p{1.17cm}
    >{\centering\arraybackslash}p{1.17cm}
    >{\centering\arraybackslash}p{1.17cm}
    >{\centering\arraybackslash}p{1.17cm}
    >{\centering\arraybackslash}p{1.17cm}
    >{\centering\arraybackslash}p{1.17cm}
    >{\centering\arraybackslash}p{1.17cm}
    >{\centering\arraybackslash}p{1.17cm}
    >{\centering\arraybackslash}p{1.17cm}
}
\Xhline{0.25ex}
\multicolumn{3}{c|}{\multirow{2}{*}{\textsc{Method}}} & \multicolumn{2}{c}{Zebra} & \multicolumn{2}{c}{PPT} & \multicolumn{2}{c}{Flower} & \multicolumn{2}{c}{Comic} & \multicolumn{2}{c}{Bird} \\
\cline{4-13}

\multicolumn{3}{c|}{}&  $\times 4$ & $\times 8$ & $\times 4$ & $\times 8$ & $\times 4$ & $\times 8$ & $\times 4$ & $\times 8$ & $\times 4$ & $\times 8$ \\
\hline
\multirow{4}{*}{\rotatebox{90}{\small \,\,\,\,\,\,Unsup\,\,\,\,\,\,\,\,}\hspace{0.6mm} \vrule}
& Bicubic & --   & 23.10 & 18.55 & 20.31 & 17.17 & 23.76 & 20.52 & 20.20 & 17.77 & 28.44 & 23.30 \\
& TV~\cite{rudin1992nonlinear} &1992    & 23.51 & 18.68 & 20.51 & 17.32 & 23.99 & 20.66 & 20.32 & 17.84 & 28.75 & 23.67 \\
& DIP~\cite{ulyanov2018deep}&2018 & 23.83 & 19.55 & 22.47 & 18.72 & 24.25 & 21.31 & 20.75 & 18.44 & 29.13 & 24.94 \\
& \cellcolor[HTML]{D7FFD7}DSP (ours)
&\cellcolor[HTML]{D7FFD7}2025
& \cellcolor[HTML]{D7FFD7}\textbf{24.31} & \cellcolor[HTML]{D7FFD7}\textbf{19.68} & \cellcolor[HTML]{D7FFD7}\textbf{22.99} & \cellcolor[HTML]{D7FFD7}\textbf{18.84} & \cellcolor[HTML]{D7FFD7}\textbf{24.57} & \cellcolor[HTML]{D7FFD7}\textbf{21.41} & \cellcolor[HTML]{D7FFD7}\textbf{20.90} & \cellcolor[HTML]{D7FFD7}\textbf{18.48} & \cellcolor[HTML]{D7FFD7}\textbf{30.27} & \cellcolor[HTML]{D7FFD7}\textbf{25.04} \\ \hdashline
\multirow{4}{*}{\rotatebox{90}{\small\,\,\,\,\,\,\,\,\,\,\,\,\,Sup\,\,\,\,\,\,\,\,\,\,\,\,\,\,\,\,}\hspace{0.6mm} \vrule}
& LapSRN~\cite{tai2017image} &2017      & 24.71 & --    & \textbf{24.26} & --    & \textbf{26.08} & --    & \textbf{21.49} & --    & {31.17} & --    \\
& ResShift~\cite{yue2023resshift}&2023  & \textbf{25.03} & --    & 23.43 & --    & 23.52 & --    & 20.86 & --    & 30.00 & --    \\
& SinSR~\cite{wang2024sinsr}&2024       & 21.51 & --    & 22.22 & --    & 22.43 & --    & 18.62 & --    & 26.55 & --    \\
& SinSR 15 &2024                   & 21.67 & --    & 22.10 & --    & 22.28 & --    & 19.09 & --    & 26.75 & --    \\
& ASID~\cite{park2025efficient} &2025
& 18.90 & --    & 22.59 & --    & 23.48 & --    & 18.90 & --    & \textbf{32.59} & -- \\
\Xhline{0.25ex}
\end{tabular}
}
\vspace{-0.1cm}
\end{table*}

\begin{table*}[t!]
\caption{Performance comparison across supervised methods (GLCIC and Prefpaint) and unsupervised methods (DIP and DSP (ours)) on inpainting tasks. Best values per image are highlighted.}
\label{inpaint_psnr_comparison}
\renewcommand{\arraystretch}{1.1}
\setlength{\tabcolsep}{4pt}
\vspace{-0.1cm}
\resizebox{\linewidth}{!}{
\hspace{0.025\linewidth}
\begin{tabular}{
    >{\centering\arraybackslash}p{0.35cm}
    >{\centering\arraybackslash}p{2.0cm}|
    >{\centering\arraybackslash}p{1.0cm}|
    >{\centering\arraybackslash}p{1.45cm}
    >{\centering\arraybackslash}p{1.45cm}
    >{\centering\arraybackslash}p{1.45cm}
    >{\centering\arraybackslash}p{1.45cm}
    >{\centering\arraybackslash}p{1.45cm}
    >{\centering\arraybackslash}p{1.5cm}
    >{\centering\arraybackslash}p{1.95cm}
    >{\centering\arraybackslash}p{1.6cm}
}
\Xhline{0.25ex}
&\multicolumn{2}{c|}{\textsc{Method}} & Vase & Street & Face & Beach & Shore & Glider& Backstreet & Library \\
\hline
\multirow{2}{*}{\rotatebox{90}{\small Unsup}\hspace{1mm} \vrule}
&DIP~\cite{ulyanov2018deep}   &2018    & 28.32 & 22.18 & 22.91 & 19.89 & 29.43 & 29.46 & 23.85 & 19.06  \\
&\cellcolor[HTML]{D7FFD7}DSP (ours) 
&\cellcolor[HTML]{D7FFD7}2025 & 
\cellcolor[HTML]{D7FFD7}\textbf{29.44} & 
\cellcolor[HTML]{D7FFD7}\textbf{22.63} & 
\cellcolor[HTML]{D7FFD7}\textbf{23.52} & 
\cellcolor[HTML]{D7FFD7}\textbf{20.06} & 
\cellcolor[HTML]{D7FFD7}\textbf{31.27} & 
\cellcolor[HTML]{D7FFD7}\textbf{29.96} & 
\cellcolor[HTML]{D7FFD7}\textbf{26.34} & 
\cellcolor[HTML]{D7FFD7}\textbf{19.37} \\  \hdashline
\multirow{2}{*}{\rotatebox{90}{\small \,\,Sup\,\,\,} \hspace{0.9mm}\vrule}&GLCIC~\cite{iizuka2017globally} & 2017& \textbf{29.51} & 18.26 & 19.49 & \textbf{20.15} & \textbf{31.12} & \textbf{30.39} & \textbf{26.77} & \textbf{18.88} \\
&Prefpaint~\cite{liu2024prefpaint}& 2024& 22.74 & \textbf{23.21} & \textbf{25.43} & 19.85 & 27.68 & 27.40 & 23.04 & 16.80\\

\Xhline{0.25ex}
\end{tabular}
}
\vspace{-0.1cm}
\end{table*}

\section{Experimental Results} 
We evaluate DSP on blind denoising, restoration, super-resolution, and inpainting tasks. Our main goal is to benchmark against DIP as the foundational implicit prior, while also comparing to recent strong unsupervised and task-specific SOTA methods. DSP is proposed with the goal of not to beat all SOTA unsupervised techniques but to introduce a new direction for frequency-domain implicit models. We report PSNR and visual results (with zoom-in views) on standard benchmarks~\citep{mairal2009non,zeyde2010single,huang2015single, ma2022regionwise, yu2019free}, using the same iteration schedule for DIP and DSP. All experiments run on a Tesla T4 GPU (15GB RAM).

\begin{figure*}[t!]
  \centering
\includegraphics[width=\textwidth]{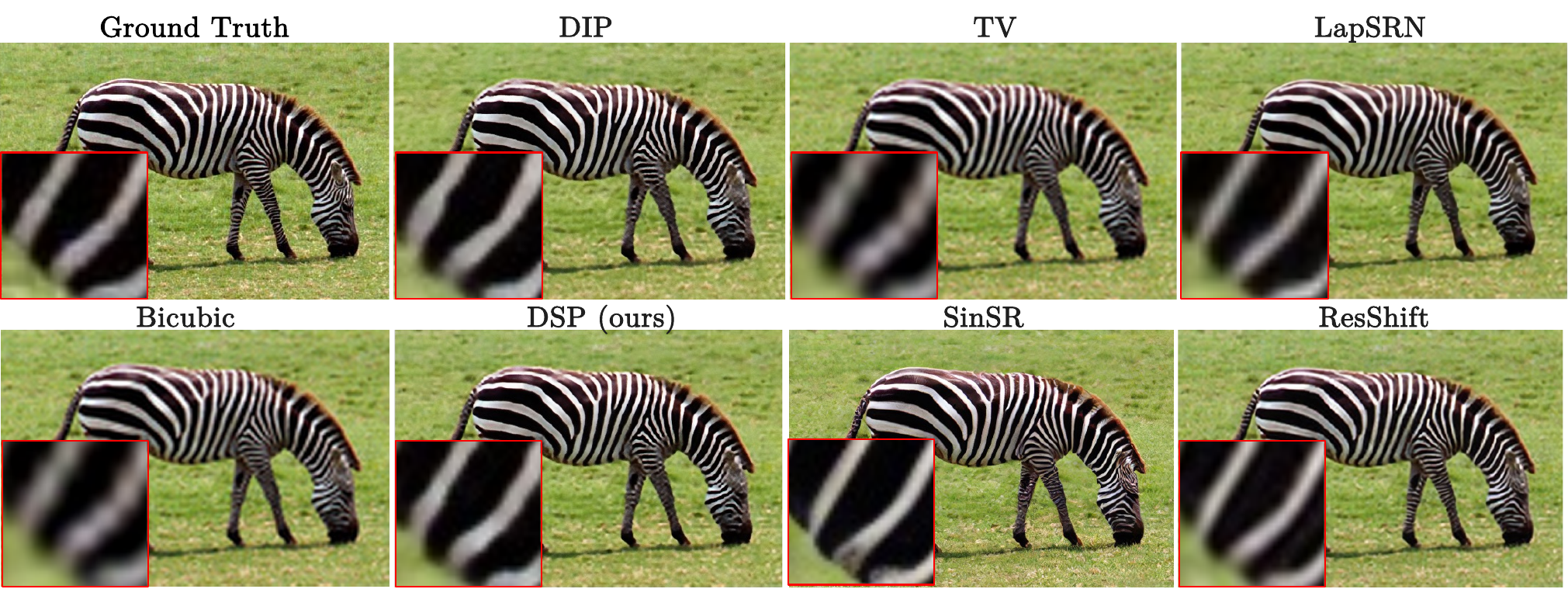}\\[-0.2cm]
        \caption{Comparison of super resolution task with 4$\times$ upscale on the ``Zebra'' image among Bicubic, TV, DIP, and the supervised methods (LapSRN, ResShift, and SinSR) with our proposed DSP. 
        }
\label{sr_zebra}
\vspace{-0.4cm}
\end{figure*}

\begin{figure*}[t!]
  \centering
\includegraphics[width=\textwidth]{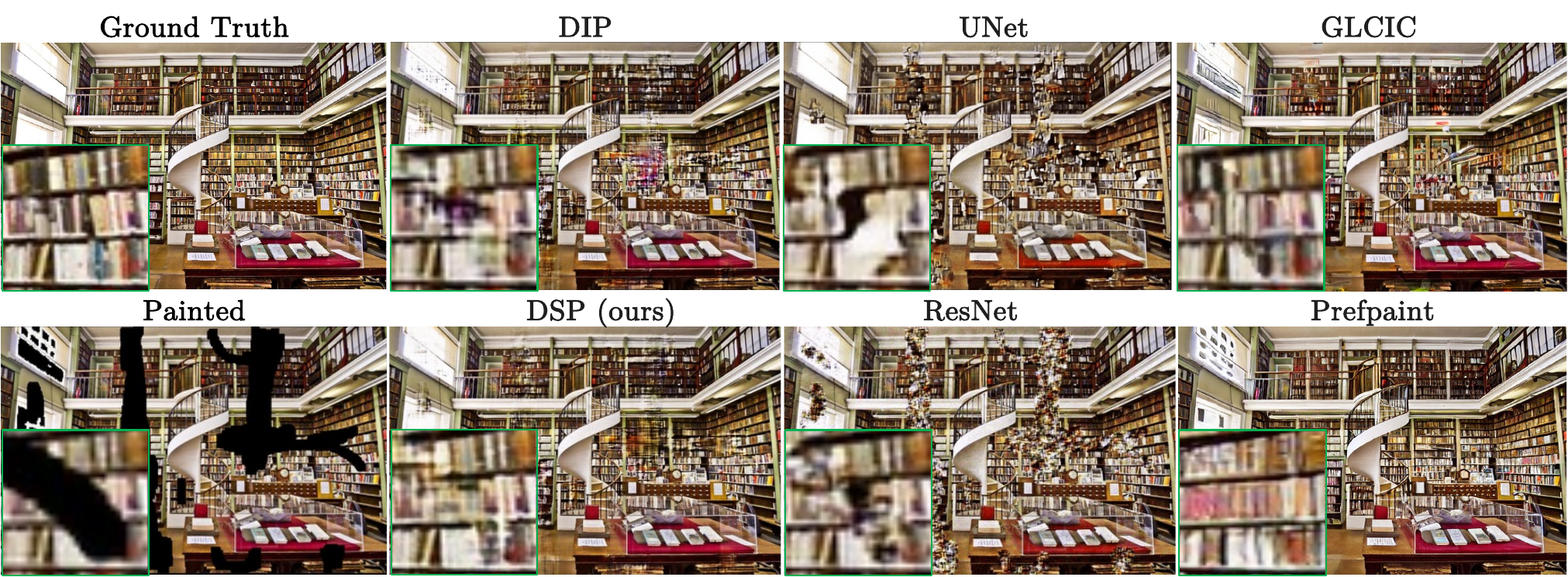}\\[-0.2cm]
    \caption{The visualisation comparison of inpainting task with on the ``Library'' image among DIP, UNet, ResNet, and GLCIC, Prefpaint, and our proposed DSP. }
\label{inpaint_library}
\vspace{-0.6cm}
\end{figure*}
\textbf{Blind Denoising.}
In the blind denoising task, the forward measurement operator is the identity operator. We compare the denoising performance (measured in PSNR, dB) of our proposed DSP method with several established unsupervised approaches, including DIP~\cite{ulyanov2018deep}, CBM3D~\cite{lebrun2012analysis}, Noise2Self~\cite{batson2019noise2self}, Wavelet-based shrinkage~\cite{chang2000adaptive}, Total Variation (TV) denoising~\cite{chambolle2004algorithm, getreuer2012rudin}, WIRE~\cite{saragadam2023wire}, and DS-N2N~\cite{bai2025dual}, across a variety of test images. As shown in Table~\ref{psnr_comparison_denoise}, DSP achieves notable improvements over the baselines, particularly on fine-textured or complex regions such as ``Plane'' and ``Brick Wall,'' with gains of +1.69\,dB and +1.53\,dB over DIP, respectively. This suggests that DIP tends to overfit to noise or overly smooth detailed textures, whereas DSP better preserves structural details. Visual comparisons in Figure~\ref{denoising_baboon} further confirm these findings: DSP delivers cleaner reconstructions with enhanced texture fidelity, avoiding the patchy artefacts commonly observed in Noise2Self and the grid-like behaviour and over-smoothing typical of CBM3D in highly structured regions.
Additionally, Figure~\ref{denoising_plane} illustrates CBM3D and Noise2Self produce visible grid-like artifacts, with Noise2Self further exhibiting noticeable over-smoothing, whereas DIP allows residual noise to persist—likely due to the iterative process gradually pulling the reconstruction closer to the noisy input—while DSP successfully preserves the sharp structure and fine details.

\vspace{0.3cm}
\textbf{Image Restoration.}
In the image restoration task, a Bernoulli mask is applied as the forward measurement operator. 
We compare our DSP method against baselines including DIP, and Convolutional Sparse Coding (CSC)~\cite{wohlberg2016boundary} implemented with~\cite{wohlberg2016sparse, wohlberg2017sporco}.
As shown in Table~\ref{psnr_comparison_restoration}, DSP consistently outperforms these methods. For example, on the ``Barbara'' image, which contains fine patterns and directional textures, DSP yields a higher PSNR than both DIP and TV, indicating better preservation of structured detail. Qualitative comparisons in Figure~\ref{restoration_barbara} further reveal that DSP reconstructs finer textures and edges more faithfully, whereas TV introduce artefacts in regions, CSC tend to produce over-smoothed results and DIP kept masked artifact with non smoothed edge features.
These results highlight the capability of DSP to recover complex structures even under aggressive masking.

\begin{figure*}[t!]
  \centering
\includegraphics[width=\textwidth]{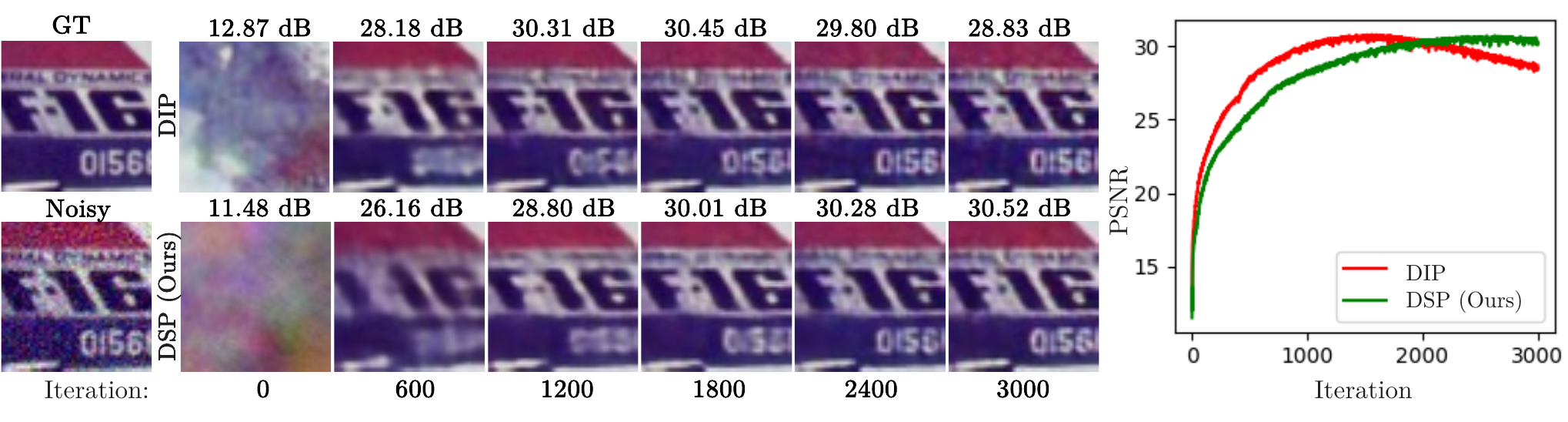}\\[-0.2cm]
  \caption{The progression over iterations of the zoomed in of blind denoising task on the ``Plane'' image, with its Peak Signal-to-Noise Ratio (PSNR) of DIP and DSP (ours) methods.}
  \label{fig:early_stop}
  \vspace{-0.3cm}
\end{figure*}

\newpage
\textbf{Super Resolution.}
For the super resolution task, the forward measurement operator corresponds to downsampling with fixed scaling factors. We compare our proposed DSP method with a range of both unsupervised and supervised baselines. The unsupervised group includes DIP~\cite{ulyanov2018deep}, Total Variation (TV)~\cite{rudin1992nonlinear}, and Bicubic upsampling, while the supervised group comprises LapSRN\cite{tai2017image}, ResShift~\cite{yue2023resshift}, SinSR~\cite{wang2024sinsr} and ASID~\cite{park2025efficient}.
Since supervised methods are trained for fixed resolutions, only the officially available $\times 4$ upscaling is evaluated.
As supervised methods are trained for fixed resolutions, only the officially available $\times 4$ results are reported. In contrast, DSP operates at arbitrary scales without any task-specific pretraining. 
As shown in Table~\ref{psnr_comparison_sr}, DSP consistently achieves the highest PSNR across all test images in both $\times 4$ and $\times 8$ settings among unsupervised methods. Notably, on the “Bird” image, DSP attains 30.27dB, outperforming DIP by over 1dB and even surpassing some supervised baselines such as SinSR and ResShift. Figure~\ref{sr_zebra} offers a visual comparison on the “Zebra” image at $\times 8$ upscaling. While supervised methods tend to produce smoother outputs, they often oversimplify textures due to reliance on generic learned priors. In contrast, DSP better preserves fine structures such as the zebra’s stripes and background foliage, showing sharper detail and fewer artefacts than DIP. These results highlight DSP’s ability to recover image-specific features and fine textures in a single-shot, unsupervised setting—making it a flexible and robust approach for real-world super resolution tasks.

\textbf{Inpainting.}
We evaluate DSP on the inpainting task using a range of unsupervised and supervised baselines. The unsupervised methods include DIP~\cite{ulyanov2018deep}, while the supervised methods are UNet~\cite{rudin1992nonlinear}, ResNet~\cite{ronneberger2015u, he2016deep}, GLCIC~\cite{iizuka2017globally} and Prefpaint~\cite{liu2024prefpaint}. 
As shown in Table~\ref{inpaint_psnr_comparison}, DSP outperforms DIP across all test images, with consistent PSNR improvements. On several images such as ``Shore'' and ``Library'', DSP even surpasses supervised methods, which confirm that DSP effectively leverages image-specific structures without requiring external training data. Qualitative results on the ``Library'' image are shown in Figure~\ref{inpaint_library}. DIP and ResNet produce dark twisted regions where the mask is present, indicating unstable reconstructions in under-constrained areas. UNet offers smoother outputs but fails to recover semantically plausible structures. GLCIC, a GAN-based method, introduces edge artefacts near mask boundaries, likely due to adversarial limitations. Although Prefpaint yields clean and realistic completions, it diverges from the ground truth content, aligning instead with learned aesthetic preferences. In contrast, DSP generates visually coherent and structurally faithful reconstructions, striking a strong balance between realism and fidelity, despite operating entirely without supervision or pretrained models.

\vspace{0.3cm}
\textbf{Early Stopping Analysis.} We analyse the iterative progression of DIP and DSP on the ``Plane'' image from the blind denoising task, as visualised in Figure~\ref{fig:early_stop}. DIP reaches its peak PSNR around 1800 iterations but then quickly dropped to 28.83 dB by iteration 3000 due to overfitting. In contrast, DSP improves steadily throughout optimisation, reaching 30.52 dB at 3000 iterations. The visual comparison shows that DSP produces increasingly cleaner and structurally consistent reconstructions, while DIP begins to introduce artefacts after its peak. This highlights DSP’s stability and robustness across iterations, especially in the absence of manual early stopping.

\section{Conclusion}
We introduced Deep Spectral Prior (DSP), a frequency-domain formulation for unsupervised image reconstruction. By shifting the optimisation objective from pixel space to the Fourier domain, DSP imposes an implicit spectral prior aligned with image statistics and the spectral bias of CNNs. This leads to sharper reconstructions, improved robustness, and removes the need for early stopping—addressing core limitations of DIP. Our theoretical analysis and empirical results jointly demonstrate that DSP offers a principled and versatile alternative to spatial-domain priors. We hope this work inspires further research into frequency-based implicit models for inverse problems.

\newpage
\section*{Acknowledgment}
YC is funded by an AstraZeneca studentship and a Google studentship.
XZ thanks Qiuzhen College, Tsinghua University, for supporting this research.
CBS acknowledges support from the Philip Leverhulme Prize, the Royal Society Wolfson Fellowship, the EPSRC advanced career fellowship EP/V029428/1, EPSRC grants EP/S026045/1 and EP/T003553/1, EP/N014588/1, EP/T017961/1, the Wellcome Innovator Awards 215733/Z/19/Z and 221633/Z/20/Z, the European Union Horizon 2020 research and innovation programme under the Marie Skodowska-Curie grant agreement No. 777826 NoMADS, the Cantab Capital Institute for the Mathematics of Information and the Alan Turing Institute. 
AIAR gratefully acknowledges the support from Yau Mathematical Sciences Center, Tsinghua University. This work is also supported by the Tsinghua University Dushi Program.

{
    \small
    \bibliographystyle{ieeenat_fullname}
    \bibliography{main}
}
\resumetoc


\newpage

\appendix
\twocolumn[
\begin{center}
{\Large{\textsc{Supplementary Material For:}\\[0.3cm] \textbf{Deep Spectral Prior}}}
\end{center}
\vspace{1cm}]
\noindent\rule{0.5\textwidth}{0.6pt}
\vspace{-0.5cm}
{\hypersetup{linkcolor=cvprblue}
\tableofcontents}
\noindent\rule{0.5\textwidth}{0.6pt}

\vspace{1cm}
\section{Supplementary Theoretical Proofs}
This supplementary section presents the complete mathematical proofs for the results stated in the main paper. These results form the theoretical foundation of the proposed methodology, and the detailed derivations provided here aim to enhance clarity. By including these extended proofs, we offer additional insight into the logical structure and assumptions underlying our approach, which may be of interest to readers seeking a deeper understanding of the theoretical aspects of our work.
\addcontentsline{toc}{subsection}{Corollary 2.1}

\vspace{0.3cm}
\noindent\textbf{Corollary 2.1}
(\textbf{Equivalence of Deep Spectral Prior Loss and Pixel-Space Error}).~\label{cor::2.1_supp}
Let \(x, \hat{x} \in \mathbb{R}^m\), and let \(\mathcal{F} : \mathbb{R}^m \to \mathbb{C}^m\) denote the unitary discrete Fourier transform. Then
\[
\|x - \hat{x}\|_2^2 = \|\mathcal{F}(x) - \mathcal{F}(\hat{x})\|_2^2.
\]
In particular, for any neural network output \(f_\theta(z)\) and observation \(y\),
$
\mathcal{L}_{\mathrm{DSP}}(\theta)
=\frac{1}{2}\big\|
\mathcal{F}(\mathcal{A} f_\theta(z)) - \mathcal{F}(y)
\big\|_2^2
$
is exactly equal to the pixel-domain reconstruction error \(\frac{1}{2}\|\mathcal{A}f_\theta(z) - y\|_2^2\). 
Thus, DSP does not alter the objective value but reformulates the optimisation in the spectral domain. 
\begin{proof}
Let $x,\hat{x} \in \mathbb{R}^{H \times W}$ be images of spatial size $H \times W$.
We use the unitary 2D discrete Fourier transform $\mathcal{F} : \mathbb{R}^{H \times W} \to \mathbb{C}^{H \times W}$.
This result follows directly from Parseval's theorem as follows:

\begin{align}
\|\mathcal{F}(x) - \mathcal{F}(\hat{x})\|_2^2
&=
\sum_{p=1}^{H}\sum_{q=1}^{W}
\left|
\mathcal{F}(x)_{p,q}-\mathcal{F}(\hat{x})_{p,q}
\right|^2.
\end{align}

\noindent
Using the unitary 2D DFT
\[
\mathcal{F}(x)_{p,q}
=
\frac{1}{\sqrt{HW}}
\sum_{m=1}^{H}\sum_{n=1}^{W}
x_{m,n}\,
e^{-2\pi i\left(\frac{pm}{H}+\frac{qn}{W}\right)},
\]
(and analogously for $\hat{x}$), the difference becomes
\begin{align}
&\|\mathcal{F}(x) - \mathcal{F}(\hat{x})\|_2^2 \nonumber\\
&=
\frac{1}{HW}
\sum_{p,q}
\left|
\sum_{m,n}
(x_{m,n}-\hat{x}_{m,n})\,
e^{-2\pi i\left(\frac{pm}{H}+\frac{qn}{W}\right)}
\right|^2 \nonumber\\
&=
\frac{1}{HW}
\sum_{p,q}\;
\sum_{m,n}
\sum_{k,\ell}
(x_{m,n}-\hat{x}_{m,n})
(x_{k,\ell}-\hat{x}_{k,\ell})^* \nonumber\\
&\qquad\qquad\times
e^{-2\pi i\left(\frac{pm}{H}+\frac{qn}{W}\right)}
e^{2\pi i\left(\frac{pk}{H}+\frac{q\ell}{W}\right)}.
\end{align}

\noindent
Using orthogonality identities
\begin{align}
    \frac{1}{H}\sum_{p=1}^{H}e^{-2\pi i\, p(m-k)/H}=\delta_{m,k}, \\
\frac{1}{W}\sum_{q=1}^{W}e^{-2\pi i\, q(n-\ell)/W}=\delta_{n,\ell},
\end{align}

\noindent
all cross-terms collapse, and we obtain
\begin{align}
\|\mathcal{F}(x) - \mathcal{F}(\hat{x})\|_2^2
&=
\sum_{m,n}|x_{m,n}-\hat{x}_{m,n}|^2 \\
&= \|x-\hat{x}\|_2^2.
\end{align}
Thus the pixel-space and Fourier-domain errors are identical.
\end{proof}
\addcontentsline{toc}{subsection}{Theorem 2.1}
\noindent\textbf{Theorem 2.1}
{\textnormal{\textbf{(Complex Descent Law: Distinct Optimisation Paths of DIP and DSP).}}}{}~\label{thm::2.1_supp}
Consider the frequency-domain loss
\begin{align}
\mathcal{L}_{\mathrm{DSP}}(\theta)
&=
\frac{1}{2}\big\|
\mathcal{F}(\mathcal{A}f_\theta(z)) - \mathcal{F}(y)
\big\|_2^2\label{eq:L2}\\
&=
\frac{1}{2}
\big(
\mathcal{F}(\mathcal{A}f_\theta(z)) - \mathcal{F}(y)
\big)^\dagger
\big(
\mathcal{F}(\mathcal{A}f_\theta(z)) - \mathcal{F}(y)
\big).\nonumber
\end{align}
where \(\mathcal{F}\) denotes the discrete Fourier transform and \({}^\dagger\) the Hermitian transpose.
The optimisation dynamics under gradient descent differ fundamentally between real-valued DIP and complex-valued DSP: \\

\vspace{0.2cm}

\begin{itemize}
\item \textbf{DIP Optimisation:} for \(\theta \in \mathbb{R}^p\),
\begin{align}\label{eq:dip_dynamics}
&    \frac{d\theta}{dt}
=
-\alpha\nabla_\theta \mathcal{L}_{\mathrm{DIP}}(\theta)\\
&=
-\alpha\,
\nabla_\theta
\big(
\mathcal{A}f_\theta(z)- y
\big)^T
\!\cdot\!
\big(
\mathcal{A}f_\theta(z) - y
\big)
,\nonumber\\
&
\text{where} \,\, \mathcal{L}_{\mathrm{DIP}}(\theta)
=
\frac{1}{2}\big\|
\mathcal{A} f_\theta(z)
-
y
\big\|_2^2\nonumber.
\end{align}

\item \textbf{DSP Optimisation:} for \(\tilde{\theta} = a + ib \in \mathbb{C}^p\),
\begin{align}
&\frac{da}{dt} = -\alpha\,\nabla_a \mathcal{L}_{\mathrm{DSP}}(\tilde{\theta}),
\frac{db}{dt} = -\alpha\,\nabla_b \mathcal{L}_{\mathrm{DSP}}(\tilde{\theta}), \nonumber\\[4pt]
&\frac{d\tilde{\theta}}{dt} = -\alpha\,\nabla_{\tilde{\theta}^*} \mathcal{L}_{\mathrm{DSP}}(\tilde{\theta}) \label{eq:dsp_dynamics}\\[4pt]
&= -\alpha\,
\nabla_{\tilde{\theta}^*}
\Big[
\big(
\mathcal{F}(\mathcal{A}f_{\tilde{\theta}}(z)) - \mathcal{F}(y)
\big)^\dagger
\!\cdot\!
\big(
\mathcal{F}(\mathcal{A}f_{\tilde{\theta}}(z)) - \mathcal{F}(y)
\big)
\Big].\nonumber
\end{align}
where \(\nabla_{\tilde{\theta}^*}\) denotes the Wirtinger gradient with respect to the complex conjugate variable.
\end{itemize}

\begin{proof}
For gradient descent method with $\theta\in \mathbb{R}^p$, the parameter $\theta$ updates as
\begin{align}
\theta^{(t+1)}=\theta^{(t)}-\alpha \frac{\partial\mathcal L}{\partial\theta}(\theta^{(t)}).
\end{align}
By extending $t$ as a continuous variable, the finite differentiation $\frac{\theta^{(t+1)}-\theta^{(t)}}{(t+1)-t}\simeq \frac{d\theta^{(t)}}{dt}+O(\alpha^2)$. And as $\alpha\rightarrow 0$, the $O(\alpha^2)$ could be ignored, and the optimisation path is approximated as 
\begin{equation}\label{contiODE}
\frac{d\theta}{dt}=-\alpha\frac{\partial\mathcal L}{\partial\theta}(\theta^{(t)}).
\end{equation}
To strictly prove this ODE approximation hold, we assume
\begin{itemize}
\item local $M$-Lipshitz condition of $\mathcal L$:
\begin{align}\|\mathcal L(u)-\mathcal L (v)\|\le M\|u-v\|.\nonumber\end{align}
\item local $K$-Lipshitz condition of $\partial_\theta \mathcal L$ :\begin{align}\|\frac{\partial\mathcal L}{\partial\theta}(u)-\frac{\partial\mathcal L}{\partial\theta}(v)\|\le K\|u-v\|. \nonumber \end{align}
\end{itemize}
Under these assumptions, the ODE’s one–step update satisfies
\begin{align}
&\hat \theta^{(t+1)}-\theta^{(t)}=-\int_{0}^{1} \alpha\partial_\theta\mathcal L(\theta^{(u+t)}) du
\nonumber\\&=-\alpha\partial_\theta\mathcal L(\theta^{(t)})-\int_{0}^{1}\alpha[\partial_\theta\mathcal L(\theta^{(u+t)})-\partial_\theta\mathcal L(\theta^{(t)}) ]du
\nonumber\\&=\frac{d\theta^{(t)}}{dt}-\int_{0}^{1}\alpha[\partial_\theta\mathcal L(\theta^{(u+t)})-\partial_\theta\mathcal L(\theta^{(t)}) ]du
\end{align}
And the discrepancy between finite differentiation and derivative is
\begin{align}
&\|\theta^{(t+1)}-\theta^{(t)}-\frac{d\theta^{(t)}}{dt}\|=\|\hat \theta^{(t+1)}-\theta^{(t)}+\alpha\partial_\theta\mathcal L(\theta^{(t)})\|\nonumber\\&\le\alpha\int_{0}^{1}\|\partial_\theta\mathcal L(\theta^{(u+t)})-\partial_\theta\mathcal L(\theta^{(t)})\|du\nonumber\\&\le\alpha K\int_{0}^{1}\|\theta^{(u+t)}-\theta^{(t)}\|du\\\nonumber&\le\alpha^2K\int_{0}^{1}\int_{0}^{u}\|\partial_\theta \mathcal L(\theta^{(s+t)})  \|dsdu\le \frac{1}{2}\alpha^2 KM\simeq O(\alpha^2)
\end{align}
Thus, the ODE approximation \eqref{contiODE} holds.  
Substituting the DIP loss into this expression yields to (\ref{eq:dip_dynamics}).

\noindent Now for our Deep Spectral Prior (DSP), the discrete gradient descent updates for DSP are
\begin{align}
&a^{(t+1)}-a^{(t)} = -\alpha\,\nabla_a \mathcal{L}_{\mathrm{DSP}}(\tilde{\theta}),\nonumber 
\\&
b^{(t+1)}-b^{(t)} = -\alpha\,\nabla_b \mathcal{L}_{\mathrm{DSP}}(\tilde{\theta}), \nonumber\\[4pt]
&\tilde{\theta}^{(t+1)}-\tilde{\theta}^{(t)} =
(a^{(t+1)}-a^{(t)})+i(b^{(t+1)}-b^{(t)})\nonumber\\&
=-\alpha\,(\nabla_a+i\nabla_b)\mathcal{L}_{\mathrm{DSP}}(\tilde{\theta})
=-2\alpha\,\nabla_{\tilde{\theta}^*} \mathcal{L}_{\mathrm{DSP}}(\tilde{\theta})
\end{align}
This shows that one needs to do gradient descent for both real part $a$ and imaginary part $b$ simultaneously in one step. 

\noindent
One can still express the procedure as ODEs, but here we should assume the $M$-Lipshitz condition for $\mathcal L_{\mathrm{DSP}} $ and $K$-Lipshitz condition for both $\nabla_{a} \mathcal L_{\mathrm{DSP}}$ and $\nabla_{b} \mathcal L_{\mathrm{DSP}}$ (i.e. $\nabla_{\tilde \theta} \mathcal L_{\mathrm{DSP}}$ and  $\nabla_{\tilde \theta^*}\mathcal L_{\mathrm{DSP}}$ ) instead.
Then
\begin{align}
&\hat a^{(t+1)}-a^{(t)}=-\int_{0}^{1} \alpha\partial_a \mathcal L(\tilde\theta^{(u+t)}) du
\nonumber\\&=-\alpha\partial_a \mathcal L(\tilde\theta^{(t)})-\int_{0}^{1}\alpha[\partial_a\mathcal L(\tilde\theta^{(u+t)})-\partial_a \mathcal L(\tilde\theta^{(t)}) ]du
\nonumber\\&=\frac{da^{(t)}}{dt}-\int_{0}^{1}\alpha[\partial_\theta\mathcal 
L(\tilde\theta^{(u+t)})-\partial_\theta\mathcal L(\tilde\theta^{(t)}) ]du
\end{align}
\vspace{-0.3cm}
\begin{align}
&\|\hat a^{(t+1)}-a^{(t)}-\frac{da}{dt}\|=\|\hat a^{(t+1)}-a^{(t)}+\alpha\partial_a\mathcal L(\tilde\theta^{(t)})\|\nonumber\\&\le\alpha\int_{0}^{1}\|\partial_a\mathcal L(\tilde\theta^{(u+t)})-\partial_a\mathcal L(\tilde\theta^{(t)})\|du\nonumber\\&\le\alpha K\int_{0}^{1}\|\tilde\theta^{(u+t)}-\tilde\theta^{(t)}\|du\nonumber\\&\le\alpha^2K\int_{0}^{1}\int_{0}^{u}\|(\partial_a+i\partial_b) \mathcal L(\theta^{(s+t)})  \|dsdu\nonumber
\\&\le \frac{\sqrt 2}{2}\alpha^2 KM\simeq O(\alpha^2)
\end{align}
The $b$ part is similar.
That's why we denote GD updates of both $a$ and $b$ as ODE.
\begin{align}
\frac{da}{dt} = -\alpha\,\nabla_a \mathcal{L}_{\mathrm{DSP}}(\tilde{\theta}),
\frac{db}{dt} = -\alpha\,\nabla_b \mathcal{L}_{\mathrm{DSP}}(\tilde{\theta}), \nonumber
\end{align}
So add the real part updates together with $i$ times that of imaginary part, we obtain
\begin{align}
&\frac{d\tilde \theta}{dt}=\frac{da}{dt} + i\frac{db}{dt}= -\alpha\,(\nabla_a \mathcal{L}_{\mathrm{DSP}}(\tilde{\theta})+i\nabla_b \mathcal{L}_{\mathrm{DSP}}(\tilde{\theta}))\nonumber\\&= -\alpha\,(\nabla_a+i\nabla_b)\mathcal{L}_{\mathrm{DSP}}(\tilde{\theta})=-2\alpha\nabla_{\tilde\theta^*}\mathcal{L}_{\mathrm{DSP}}(\tilde{\theta})\\\nonumber&
=-\alpha\,
\nabla_{\tilde{\theta}^*}
\Big[
\big(
\mathcal{F}(\mathcal{A}f_{\tilde{\theta}}(z)) - \mathcal{F}(y)
\big)^\dagger
\!\cdot\!
\big(
\mathcal{F}(\mathcal{A}f_{\tilde{\theta}}(z)) - \mathcal{F}(y)
\big)
\Big].
\end{align}
\end{proof}
\addcontentsline{toc}{subsection}{Theorem 2.2}
\noindent\textbf{Theorem 2.2}
{\textnormal{\textbf{(Robust Stopping Property of  DSP).}}}{}~\label{thm::2.2_supp}
For a DSP optimisation trajectory, the {NTK approximation} takes the form
\begin{equation}\label{NTK_approx_spectral}
\mathcal{F}(\mathcal A f_{\tilde\theta}(z))
=
\mathcal{F}(\mathcal Af_{\tilde{\theta}_0}(z)) + \tilde{\Phi}(z)\tilde{u},
\end{equation}
where $\tilde{\Phi}(z)=\frac{\partial \mathcal F(\mathcal{A}f_{\tilde{\theta}}(z))}{\partial \tilde\theta}\in\mathbb{C}^{N\times p}$ is assumed to be a constant over steps, with $N = H\times W\times3$, $p=\dim(\tilde \theta)$, and parameters without initialisation $\tilde u=\tilde\theta-\tilde\theta_0$ . 
DSP optimisation path (\ref{eq:dsp_dynamics}) guides parameters $\tilde u$ to update along the path:
\begin{equation}\label{eq:DSP-parameter}
\frac{d\tilde u}{dt}=
-\alpha\,\tilde{\Phi}(z)^\dagger(\tilde{\Phi}(z)\tilde{u} - \tilde{y}),
\end{equation}
where $\alpha$ is the gradient descent step size. 
The evolution of the discrepancy along gradient descent follows a linear ordinary differential equation (ODE) governed by the Hermitian operator $\tilde{\Phi}\tilde{\Phi}^\dagger$:
\begin{equation}\label{eq:DSP-picture}
\frac{d(\tilde{\Phi}(z)\tilde{u} - \tilde{y})}{dt}
=
-\alpha\,\tilde{\Phi}(z)\tilde{\Phi}(z)^\dagger(\tilde{\Phi}(z)\tilde{u} - \tilde{y}),
\end{equation} 
The ODE admits the closed-form solution
\begin{equation}\label{eq:DSP_solution}
\mathcal{F}(\mathcal A f_{\tilde\theta^{(t)}}(z))
=
\sum_{k=1}^{N}(1 - e^{-\alpha\mu_k t})
\langle \tilde{y}, \tilde{\psi}_k\rangle \tilde{\psi}_k,
\end{equation}
where $\tilde{\psi}_k$ and $\mu_k$ satisfy the eigendecomposition
$\tilde{\Phi}\tilde{\Phi}^\dagger
=
\sum_{k=1}^N \mu_k \tilde{\psi}_k \tilde{\psi}_k^\dagger$.
The mean-squared error (MSE) along the optimisation path of DSP evolves as
\begin{align}
&MSE_t
= \mathbb{E}\Vert \mathcal{F}(\mathcal A f_{\tilde{\theta}^{(t)}}(z)) - \tilde x\Vert^2 \label{eq:mse_t}\\
&= \sum_{k=1}^{N}\!\Big[
 (1 - e^{-\alpha\mu_k t})^2(\mathbb{E}|\langle \tilde{\epsilon}, \tilde{\psi}_k\rangle|^2+|\langle\tilde{\mathcal A x},\tilde\psi_k\rangle|^2)\nonumber\\
 & -
 2(1 - e^{-\alpha\mu_k t})Re\big
 \{\langle\tilde{\mathcal Ax},  \tilde\psi_k\rangle^*\langle\tilde x,  \tilde\psi_k\rangle\big\}+|\langle \tilde x,\tilde \psi_k \rangle|^2
\Big]\!. \nonumber
\end{align}
where $\tilde{x}=\mathcal F(x)$, $\tilde{\mathcal A x}=\mathcal F(\mathcal {A}x)$ and noise $\tilde \epsilon=\mathcal F(\epsilon)$.
The optimal stopping time for each mode $k$ is then
\vspace{-0.2cm}
\begin{align}
t_k^* &= \frac{1}{\alpha\mu_k}\ln\!\left(\rho_k\right), \label{eq:DSP_stopping} \\
\rho_k &=
\frac{\mathbb{E}|\langle \tilde{\epsilon}, \tilde{\psi}_k\rangle|^2+|\langle\tilde{\mathcal A x},\tilde\psi_k\rangle|^2}
{\mathbb{E}|\langle \tilde{\epsilon}, \tilde{\psi}_k\rangle|^2+|\langle\tilde{\mathcal A x},\tilde\psi_k\rangle|^2
- Re\{\langle\tilde {\mathcal{A}x},\tilde\psi_k\rangle^*\langle\tilde x,\tilde \psi_k\rangle\}}. \notag
\end{align}

\noindent Different eigenvalues $\mu_k$ induce distinct convergence speeds, endowing DSP with an inherent robustness to  stopping time. (See proof in the Supplementary Material)
\vspace{0.2cm}
\begin{proof}
For a DSP optimisation path, use a linear approximation to the network 
\begin{align}
&\mathcal F(\mathcal{A}f_{\tilde\theta}(z))=\nonumber\\&
\mathcal F(\mathcal{A}f_{\tilde\theta_0}(z))+ \nabla_{\tilde\theta}\mathcal F( \mathcal{A}f_{\tilde\theta_0}(z))(\tilde\theta-\tilde\theta_0)+O(\Vert\tilde\theta-\tilde\theta_0\Vert^2),
\end{align}
denote the $\nabla_\theta\mathcal F( A f_{\theta_0}(z))$ as a constant matrix $\tilde \Phi_{n,k}$,where $k$ is index for parameter $\theta$, $n$ is index for spectral of output image, and denote $\tilde u=\tilde \theta-\tilde \theta_0\in \mathbb{C}^p$. In here $\mathcal F(\mathcal A f_{\tilde \theta}(z))$ is complex-analytic w.r.t. $\tilde \theta$. Then we approximate the network as Neural Tanget Kernel version
\begin{equation}\label{eq:NTK_approx_spectral}
\mathcal{F}(\mathcal A f_{\tilde\theta}(z))
=\mathcal{F}(\mathcal Af_{\tilde{\theta}_0}(z)) + \tilde{\Phi}(z)\tilde{u}.
\end{equation} 
This approximation holds with
period updates as $O(\|\tilde \theta-\tilde\theta_0\|^2)\rightarrow0$.
\\
The loss function of \textbf{Deep Spectral Prior}(DSP) is approximated as 
\begin{equation}
	\begin{split}
	&\mathcal{L}_{2}(\tilde u, \tilde u^*)=\frac{1}{2}\Vert \mathcal F (\mathcal {A} f_{\tilde\theta}(z))-\mathcal F (y)\Vert^2=
	\\&\frac{1}{2}\Vert \mathcal F( \mathcal{ A} f_{\tilde\theta_0}(z))+\tilde\Phi(z)\tilde u-\mathcal F(y) \Vert^2=
	\\&\frac{1}{2}[(\mathcal F( \mathcal{A}f_{\tilde\theta_0}(z))+\tilde\Phi(z)\tilde u-\mathcal F(y))^\dagger\\&(\mathcal F(\mathcal{A} f_{\tilde\theta_0}(z))+\tilde\Phi(z)\tilde u-\mathcal F(y))]
	\end{split}
\end{equation} 
First, the loss function now becomes $\mathcal {L}_{\mathrm{DSP}}(\tilde \theta)\simeq\mathcal{L}_2(\tilde u)$.
Our parameter is reparametrized as $\tilde u=u+iv$ ( where $\tilde \theta_0=a_0+ib_0$, $\tilde \theta=a+ib$, $\tilde u=\tilde \theta-\tilde\theta_0$). 
\\
Then, imposing a gradient descent on both $a$ and $b$ is exactly imposing a gradient descent on $u$  and $v$. The gradient descent becomes ODEs
\begin{subequations}\label{eq:NTK-uvpara}
	\begin{align}
		&\frac{du}{dt}=-\alpha \frac{\partial \mathcal{L}_2(\tilde u,\tilde u^*) }{\partial u}\nonumber\\&=-\frac{\alpha}{2}[\tilde \Phi(z) ^\dagger (\tilde \Phi(z)\cdot (u+iv)+\mathcal F(\mathcal{A} f_{\tilde\theta_0}(z))-\tilde y)\nonumber\\&+\tilde \Phi(z) ^T (\tilde \Phi(z)^*\cdot (u-iv)+\mathcal F(\mathcal{A} f_{\tilde\theta_0}(z))^*-\tilde y^*)]\\&\frac{dv}{dt}=-\alpha \frac{\partial \mathcal{L}_2(\tilde u,\tilde u^*) }{\partial v}\nonumber\\&
        =-\frac{\alpha}{2}[(-i)\tilde \Phi(z) ^\dagger (\tilde \Phi(z)\cdot (u+iv)+\mathcal F(\mathcal{A} f_{\tilde\theta_0}(z))-\tilde y)\nonumber\\&+i\tilde \Phi(z) ^T (\tilde \Phi(z)^*\cdot (u-iv)+\mathcal F(\mathcal{A} f_{\tilde\theta_0}(z))^*-\tilde y^*)]
	\end{align}
\end{subequations}
Here we can check the two Lipshitz conditions to obtain $O(a^2)$ accuracy under NTK.
\begin{align}
&\|\frac{\partial \mathcal{L}_2(\tilde u,\tilde u^*) }{\partial a}\|\le \|Re(\tilde \Phi^\dagger(\mathcal {F}(\mathcal{A}f_{\tilde \theta_0})-\tilde y))\|+\|\tilde\Phi\|_2^2\|\tilde u\|\nonumber\\&
\|\frac{\partial \mathcal{L}_2(\tilde u,\tilde u^*) }{\partial b}\|\le \|Re(\tilde \Phi^\dagger(\mathcal {F}(\mathcal{A}f_{\tilde \theta_0})-\tilde y))\|+\|\tilde\Phi\|_2^2\|\tilde u\|\nonumber
\end{align}
Since $\tilde u $ should be a bounded value due to the good initialisation required by NTK, the upper bounds of partial derivative on RHS should be controlled by an
\begin{align}M\le 2\|Re(\tilde \Phi^\dagger(\mathcal {F}(\mathcal{A}f_{\tilde \theta_0})-\tilde y))\|+2\|\tilde\Phi\|_2^2\max_{\tilde u}\|\tilde u\|\end{align}
And by triangle inequality
\begin{align}
&|\mathcal L_2(\tilde u,\tilde u^*)-\mathcal L_2(\tilde w,\tilde w^*)|\nonumber\\&
=|\mathcal L_2(p+iq,p-iq)-\mathcal L_2(p'+iq',p'-iq')|
\nonumber\\&\le\|\int_p^{p'}\partial_a\mathcal{L}_2 da\|+\|\int_q^{q'}\partial_b\mathcal{L}_2 db\|\le M\|\tilde u-\tilde w\|
\end{align}
Thus, bounded gradient condition i.e. $M$-Lipshitz condition is satisfied.
Besides, by 
\begin{align}
&\|\partial_a\mathcal{L}_2(\tilde u,\tilde u^*)-\partial_a\mathcal{L}_2(\tilde w,\tilde w^*)\|\nonumber\\&=\|\tilde\Phi^\dagger\tilde\Phi(\tilde u-\tilde w)+\tilde\Phi^T\tilde\Phi^*(\tilde u-\tilde w)^*\|\le 2\|\tilde \Phi\|_2^2\|\tilde u-\tilde w\|\\
&\|\partial_b\mathcal{L}_2(\tilde u,\tilde u^*)-\partial_b\mathcal{L}_2(\tilde w,\tilde w^*)\|\nonumber\\&=\|\tilde\Phi^\dagger\tilde\Phi(\tilde u-\tilde w)-\tilde\Phi^T\tilde\Phi^*(\tilde u-\tilde w)^*\|\le 2\|\tilde \Phi\|_2^2\|\tilde u-\tilde w\|,
\end{align}
The $K$-Lipshitz condition is satisfied for a $K\le2\|\Phi\|_2^2 $.
Since the two sets of Lipshitz condition is satisfied, the ODEs (\ref{eq:NTK-uvpara}) could be approximation to gradient descent up to $O(\alpha^2)$.
\par
For (\ref{eq:NTK-uvpara}), add real part with $i$ times imaginary part and obtain
\begin{equation}
	\frac{d(u+iv)}{dt}=-\alpha\tilde \Phi ^\dagger (\tilde \Phi (u+iv)-\tilde y).
\end{equation}
Multiply both sides with $\tilde \Phi$ the ODE for cDSP is now
\begin{equation}
	\frac{d (\tilde \Phi(z) \tilde u-\tilde y)}{dt}=-\alpha\tilde \Phi(z)\tilde\Phi(z)^\dagger(\tilde \Phi(z)\tilde u-\tilde y).
\end{equation}
Since Hermitian matrix have an unitary diagonalisation, we can solve the ODE as
	\begin{equation}
		\mathcal F(\mathcal{A}f_{\theta^{(t)}}(z))=\Sigma_{k=1}^{N}(1- e^{-\alpha\mu_k t})\langle \tilde y,\tilde\psi_k\rangle \tilde\psi_k
	\end{equation}
	where $\psi_k$ and $\mu_k$ satisfy $\tilde\Phi\tilde\Phi^\dagger=\Sigma_{k=1}^N\mu_k\tilde\psi_k\tilde\psi_k^\dagger$
	and $\alpha$ is the step size constant of GD.
\par
The MSE changing along the optimisation path of cDSP is obtained as
\begin{align}
		MSE_t&=\mathbb{E}\Vert \mathcal F(\mathcal {A}f_{\theta^{(t)}}(z))-\tilde x\Vert^2\nonumber\\&=\mathbb{E}\|\Sigma_{k=1}^{N}(1- e^{-\alpha\mu_k t})\langle \tilde y,\tilde\psi_k\rangle \tilde\psi_k-\langle\tilde x,\tilde\psi_k\rangle\tilde\psi_k\|^2
        \nonumber\\&
        =\mathbb{E}\Sigma_{k=1}^N|(1- e^{-\alpha\mu_k t})\langle \tilde y,\tilde\psi_k\rangle-\langle\tilde x,\tilde\psi_k\rangle|^2
        \nonumber\\&=\Sigma_{k=1}^N\mathbb{E}|(1- e^{-\alpha\mu_k t})\langle \tilde y,\tilde\psi_k\rangle-\langle\tilde x,\tilde\psi_k\rangle|^2
\end{align}
Insert $\tilde y=\tilde{\mathcal A x}+\tilde\epsilon$, expand the terms and discard the cross term with $\tilde \epsilon$ for the noise $\mathbb{E}\tilde\eta=0$, we obtain
\begin{align}
       & MSE_t=\Sigma_{k=1}^N\mathbb{E}|(1- e^{-\alpha\mu_k t})\langle \tilde {\mathcal A x}+\tilde\epsilon,\tilde\psi_k\rangle-\langle\tilde x,\tilde\psi_k\rangle|^2\nonumber\\&=\Sigma_{k=1}^{N}\big[|(1- e^{-\alpha\mu_k t})\langle \tilde{\mathcal{A} x},\tilde\psi_k\rangle-\langle\tilde{ x},\tilde\psi_k\rangle|^2\nonumber\\&+(1-e^{-\alpha\mu_k t})^2\mathbb{E}|\langle\tilde \epsilon,\tilde\psi_k\rangle|^2\big]
        \nonumber\\&=\Sigma_{k=1}^{N}\bigg[|\langle \tilde x, \tilde\psi_k \rangle|^2-2(1-e^{-\alpha\mu_k t})\mathrm{Re}\big\{\langle\tilde {\mathcal A x},\tilde\psi_k\rangle^*\langle \tilde x,\tilde\psi_k\rangle\big\}
        \nonumber\\&+(1-e^{-\alpha\mu_k t})^2(|\langle \tilde{\mathcal{A} x},\tilde\psi_k\rangle|^2+\mathbb{E}|\langle\tilde \eta,\tilde\psi_k\rangle|^2)\bigg]
\end{align} 
\par
For each channel $k\in[N]$, its contribution to the loss is 
\begin{align}
	&MSE(k)_t=\Sigma_{k=1}^{N}\bigg[|\langle \tilde x, \tilde\psi_k \rangle|^2-2\beta_k\mathrm{Re}\big\{\langle\tilde {\mathcal A x},\tilde\psi_k\rangle^*\langle \tilde x,\tilde\psi_k\rangle\big\}
        \nonumber\\&+\beta_k^2(|\langle \tilde{\mathcal{A} x},\tilde\psi_k\rangle|^2+\mathbb{E}|\langle\tilde \eta,\tilde\psi_k\rangle|^2)\bigg]\nonumber\\&       
        \beta_k=1-e^{-\alpha\mu_k t}
\end{align}
By imposing $\frac{dMSE(k)_t}{dt}=0$, the minimal MSE time i.e. optimal stopping time for $k$ th channel is
	\begin{equation}
		\beta_k^*=\frac{\mathrm{Re}\big\{\langle\tilde {\mathcal A x},\tilde\psi_k\rangle^*\langle \tilde x,\tilde\psi_k\rangle\big\}}{|\langle \tilde{\mathcal{A} x},\tilde\psi_k\rangle|^2+\mathbb{E}|\langle\tilde \eta,\tilde\psi_k\rangle|^2}
	\end{equation}
And by $t_k^*=\frac{1}{\alpha\mu_k}\ln(\frac{1}{1-\beta_k^*})$, we obtain (\ref{eq:DSP_stopping}) as
\begin{align}
t_k^* &= \frac{1}{\alpha\mu_k}\ln\!\left(\rho_k\right), \\
\rho_k &=
\frac{\mathbb{E}|\langle \tilde{\epsilon}, \tilde{\psi}_k\rangle|^2+|\langle\tilde{\mathcal A x},\tilde\psi_k\rangle|^2}
{\mathbb{E}|\langle \tilde{\epsilon}, \tilde{\psi}_k\rangle|^2+|\langle\tilde{\mathcal A x},\tilde\psi_k\rangle|^2
- Re\{\langle\tilde {\mathcal{A}x},\tilde\psi_k\rangle^*\langle\tilde x,\tilde \psi_k\rangle\}}. \notag
\end{align}
\end{proof}
\vspace{0.2cm}

\section{Supplementary Visualisations}
This figure presents supplementary visual comparisons for a variety of inverse problems, including image denoising, deblurring, inpainting, and super-resolution. Each column pair displays the degraded (before) and reconstructed (after) image results. The examples include diverse content demonstrating the robustness and generalisation capability of the proposed method.
\newpage
\begin{figure*}[ht]
\centering
\label{appendix-teaser}
  \centering
\includegraphics[width=0.98\textwidth]{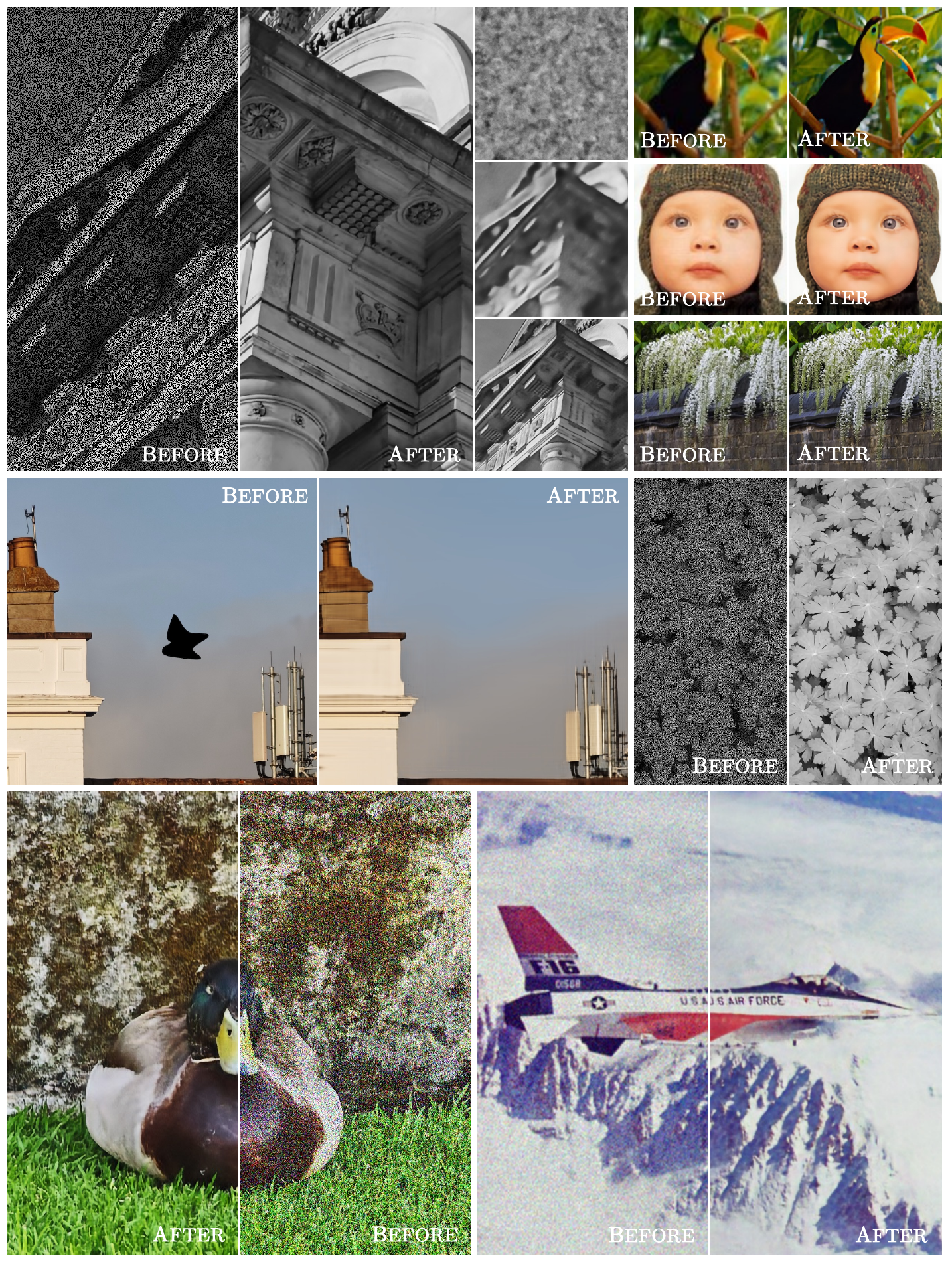}
\end{figure*}

\end{document}